%% file: main.tex
\documentclass[sigconf]{acmart}

\AtBeginDocument{%
  }

\setcopyright{none}
\renewcommand\footnotetextcopyrightpermission[1]{}
\acmConference[KDD '27]{The 33rd ACM SIGKDD Conference on Knowledge Discovery and Data Mining}{August 1--5, 2027}{San Jose, CA, USA}
\acmYear{2027}
\copyrightyear{2027}
\acmDOI{}
\acmISBN{}

\usepackage{booktabs}
\usepackage{multirow}
\usepackage{array}
\usepackage{graphicx}
\usepackage[percent]{overpic}
\usepackage{placeins}
\usepackage{pifont}
\usepackage{enumitem}
\usepackage{needspace}
\usepackage[most]{tcolorbox}   

\newtcolorbox{newadd}{%
  breakable, enhanced,
  colback=blue!8,              
  colframe=blue!45!black,      
  boxrule=0.5pt, arc=1.5pt,
  left=6pt, right=6pt, top=6pt, bottom=6pt}
  
\newcommand{\cmark}{\ding{51}}
\newcommand{\xmark}{\ding{55}}
\newcommand{\pmark}{\raisebox{0.15ex}{\scriptsize$\triangle$}}
\definecolor{figNavy}{HTML}{0B1F3A}
\definecolor{figTeal}{HTML}{147D7E}
\definecolor{figOrange}{HTML}{E76F2E}

\providecommand{\experimentsappendix}{}

\begin{document}

\title{OfficePrism: Benchmarking Agents on Single-Domain Operations Across Enterprise Business Functions}

\author{Hanfeng Liu}
\authornote{Equal Contribution.}
\affiliation{%
  \institution{Alibaba Group}
  \city{Hangzhou}
  \country{China}}
\affiliation{%
  \institution{HKUST(GZ)}
  \city{Guangzhou}
  \country{China}}
\email{hliu174@connect.hkust-gz.edu.cn}

\author{Zhenyu Luo}
\authornotemark[1]
\affiliation{%
  \institution{Alibaba Group}
  \city{Hangzhou}
  \country{China}}
\affiliation{%
  \institution{HKUST(GZ)}
  \city{Guangzhou}
  \country{China}}
\email{zluo842@connect.hkust-gz.edu.cn}

\author{Dongchao Zhang}
\affiliation{%
  \institution{Alibaba Group}
  \city{Hangzhou}
  \country{China}}
\affiliation{%
  \institution{HKUST(GZ)}
  \city{Guangzhou}
  \country{China}}
\email{dzhang223@connect.hkust-gz.edu.cn}

\author{Zexi Liu}
\affiliation{%
  \institution{Alibaba Group}
  \city{Hangzhou}
  \country{China}}
\affiliation{%
  \institution{HKUST(GZ)}
  \city{Guangzhou}
  \country{China}}
\email{zliu403@connect.hkust-gz.edu.cn}

\author{Zulong Chen}
\affiliation{%
  \institution{Alibaba Group}
  \city{Hangzhou}
  \country{China}}
\email{zulong.czl@alibaba-inc.com}

\author{Jinxin Hu}
\affiliation{%
  \institution{Alibaba International Digital Commerce Group}
  \city{Beijing}
  \country{China}}
\email{jinxin.hjx@lazada.com}

\author{Jia Xu}
\affiliation{%
  \institution{Cyberspace Institute of Advanced Technology, Guangzhou University}
  \city{Guangzhou}
  \country{China}}
\email{xujia@gzhu.edu.cn}

\author{Zeyi Wen}
\authornote{Corresponding author.\\This work was done while Hanfeng Liu worked at Alibaba Group.\\Code and data are available at \url{https://github.com/Leozyu/OfficePrism}.}
\affiliation{%
  \institution{HKUST(GZ)}
  \city{Guangzhou}
  \country{China}}
\email{wenzeyi@hkust-gz.edu.cn}


\makeatletter
\gdef\addresses{%
  \@author{%
    \large
    \begin{tabular}{@{}c@{}}
      Hanfeng Liu$^{1,2,*}$\quad
      Zhenyu Luo$^{1,2,*}$\quad
      Dongchao Zhang\textsuperscript{1,2}\quad
      Zexi Liu\textsuperscript{1,2}\\[0.35ex]
      Zulong Chen$^{1,\dagger}$\quad
      Jinxin Hu\textsuperscript{3}\quad
      Jia Xu\textsuperscript{4}\quad
      Zeyi Wen$^{2,\dagger}$
    \end{tabular}}%
  \affiliation{obeypunctuation=true}{%
    \institution{%
      \small
      \begin{tabular}{@{}c@{}}
        \textsuperscript{1}Alibaba Group, Hangzhou, China\quad
        \textsuperscript{2}HKUST(GZ), Guangzhou, China\\
        \textsuperscript{3}Alibaba International Digital Commerce Group, Beijing, China\\
        \textsuperscript{4}Cyberspace Institute of Advanced Technology, Guangzhou University, Guangzhou, China\\[0.25ex]
        zulong.czl@alibaba-inc.com, 
        wenzeyi@hkust-gz.edu.cn\quad
      \end{tabular}}%
    \city{}%
    \country{}}}
\makeatother

\begin{abstract}
Broad workplace scores can mix competence within one business function with
the extra cost of coordinating across functions and systems, making failures
hard to localize. \textbf{OfficePrism} evaluates this prerequisite through
420 single-domain enterprise tasks spanning seven business functions under
shared, benchmark-defined L1--L3 task-structure levels. Each task runs in a
fresh environment combining a shared synthetic organization with one
function-specific domain and typed state access. Our pipeline uses grounded
records, business rules, and tool schemas to jointly construct a Chinese-language
ticket, a hidden executable reference witness, and an expected response,
followed by automated filtering and human curation. The same task set is
evaluated through full-schema and search-and-dispatch packages. Across 124
model--harness--package configurations, the per-case joint
natural-language-and-database pass rate ranges from 22.6\% to 76.4\%, with a
median of 67.9\%. L1 is highest in all 124 configurations, and L2 exceeds L3 in
122. Among 50 configurations sharing the same harness and model labels across
both packages, search scores higher in 30, ties in one, and scores lower in 19.
\end{abstract}

\ccsdesc[500]{Computing methodologies~Artificial intelligence}
\ccsdesc[300]{Information systems~Information systems applications}

\keywords{language agents, enterprise workflows, tool use, benchmark construction, executable evaluation}

\maketitle

\input{sections/introduction}
\input{sections/related_work}
\input{sections/benchmark}
\input{sections/construction}
\input{sections/evaluation}
\input{\experimentssection}
\input{sections/limitations_conclusion}
\input{sections/acknowledgement}
\bibliographystyle{ACM-Reference-Format}
\bibliography{references}

\appendix
\input{sections/appendix}
\experimentsappendix

\end{document}

%% file: sections/introduction.tex
\section{Introduction}
\label{sec:introduction}

Broad workplace evaluation can obscure a prerequisite for capable office
agents. A single score may mix application navigation, workspace management,
project coordination, and operations within a business function, making
failures difficult to localize. Consider an employee request to identify the
head of a department, find that person's primary email account, and enable an
out-of-office message. The operation is closed within the shared organization
plus the information-technology function, yet it already requires an
organizational lookup, a dependent record lookup, and a state-changing call.
Before asking agents to orchestrate across applications or business functions,
we therefore need to measure whether they can complete operations closed
within one function and how performance varies across benchmark-defined
levels of within-function task structure.

Prior work has established much of the required foundation. Stateful
enterprise benchmarks provide outcome and state evidence for tool execution
~\citep{styles2024workbench,styles2026workbenchrevisited,
malay2026enterpriseopsgym,pysklo2026agentdiff}. Grounded construction
pipelines derive tasks from organizational context, schemas, executable
trajectories, or shared specifications
~\citep{vishwakarma2025enterprisebench,mo2026entworld,
agarwal2026enterpriselab,ivanov2026anchor}. Office-application benchmarks
also cover long-horizon multi-application workflows
~\citep{wang2024officebench,wang2025odysseybench}. The remaining need is not
another executable enterprise world or another scalable construction pipeline.
OfficePrism complements these benchmarks by confining each task to one business
function and applying the same task-structure definition within every function.
This design supports localized evaluation before broader multi-function
coordination.

We introduce \textbf{OfficePrism}, a single-domain benchmark for
single-domain enterprise operations. Each task is closed within a task-local
environment that combines a shared synthetic organization with exactly one of
seven business functions and accesses state through typed tools. The same
task-structure criteria define L1--L3 within every function. The benchmark
artifact, OfficePrism-420, contains 420 tasks and will be released publicly
upon acceptance.

A state-grounded construction pipeline creates the task artifact. From one
grounding packet of caller context, business rules, tool schemas, and records,
the generator jointly proposes an agent-facing Chinese ticket and two hidden
annotations: a reference witness and an expected response. Replaying the
witness checks that its functions and arguments execute in the task
environment. Quality filtering, redundancy control, and human curation then
screen candidates before benchmark-set selection.

We evaluate the same 420 tasks through two interface packages.
\texttt{full} exposes native function schemas, whereas \texttt{search}
provides discovery and dispatch over the same underlying toolkit.
Section~\ref{sec:experiments} reports results across 124
model--harness--package configurations.

Our contributions are:
\begin{itemize}[leftmargin=*,nosep]
    \item \textbf{A function-resolved benchmark artifact.}
    OfficePrism-420 comprises 420 function-closed tasks across seven business
    functions under a shared L1--L3 definition.
    \item \textbf{A state-grounded construction pipeline.}
    A shared grounding packet couples an agent-facing ticket with a hidden
    witness and expected response, followed by executable replay, quality
    filtering, redundancy control, and human curation.
    \item \textbf{An empirical evaluation across 124 configurations.}
    Joint natural-language-and-database pass rates range from 22.6\% to
    76.4\%; L1 is highest in every configuration, and the full/search ordering
    varies across models and harnesses.
\end{itemize}

%% file: sections/related_work.tex
\section{Related Work}
\label{sec:related}

\noindent\textit{Office and workplace agent benchmarks.}
OfficeBench and OdysseyBench emphasize multi-application execution~\citep{wang2024officebench,wang2025odysseybench}. Others center platforms, simulated firms, or workplace sessions~\citep{drouin2024workarena,boisvert2024workarenaplus,xu2025agentcompany,zhong2026enterpriseclawbench}, while Workspace-Bench centers file workspaces~\citep{tang2026workspacebench}. Workspace-Bench structures tasks by file dependencies, EnterpriseBench by domains and subtasks, EntWorld by complexity scores, and ERP-Bench by parametric regimes~\citep{tang2026workspacebench,vishwakarma2025enterprisebench,mo2026entworld,ivanov2026anchor}. Scores spanning functions can mix within-function competence with coordination cost. Among these benchmarks, EntWorld is the closest structural neighbor because it combines system-specific tasks, difficulty scores, and SQL checks, but its unit is software system rather than business function. OfficePrism instead uses the business function as its task unit.

Function means an organizational responsibility such as HR or finance, not an API. Every task is closed within one separately loaded function-specific domain under common L1--L3 definitions.

Broader agent benchmarks cover text environments, websites, visual interfaces,
desktop operating systems, mobile devices, and multi-turn diagnostics
~\citep{liu2024agentbench,mialon2024gaia,zhou2024webarena,
koh2024visualwebarena,xie2024osworld,rawles2025androidworld,
ma2024agentboard}. API-centered suites instead stress tool selection,
multi-step execution, interaction, and stability as tool catalogs grow
~\citep{li-etal-2023-api,qin2024toolllm,wang2024gta,wang2024mint,
guo2024stabletoolbench,huang2024ultratool,patil2025bfcl,
lu2025toolsandbox,shen2025shortcutsbench,elder2026liveapi}.
These benchmarks establish the difficulty of interaction and interface scale,
but their organizing unit is an environment, application, or API collection
rather than an enterprise business function.

Table~\ref{tab:related-positioning} compares organizing units, single-function
task scope, and executable validation. Difficulty remains in prose because
definitions differ. App, WS, Ops, and Fn denote application, worker workspace,
enterprise operation, and business function. Filled and hollow dots mark
primary and non-primary units. Fn requires organizational responsibility rather
than an application or toolkit label. A 1-Fn check marks task sets whose
records, tools, and workflow remain within one function. Triangles mean partial
or variant support, and dashes no documentation. Build denotes construction
replay; State denotes programmatic post-run checking of environment or artifact
state.

\begingroup
\renewcommand{\cmark}{\textcolor{figTeal}{\ding{51}}}
\renewcommand{\pmark}{\textcolor{figOrange}{\raisebox{0.15ex}{\scriptsize$\triangle$}}}
\newcommand{\umark}{\textcolor{figNavy}{\textbullet}}
\newcommand{\uoff}{\textcolor{gray}{\raisebox{0.05ex}{$\circ$}}}
\newcommand{\nmark}{\textcolor{gray}{--}}
\begin{table}[htb]
\caption{Selected benchmarks by organizing unit, single-function task scope, and executable verification.}
\label{tab:related-positioning}
\centering
\footnotesize
\setlength{\tabcolsep}{1.2pt}
\renewcommand{\arraystretch}{1.04}
\begin{tabular*}{\columnwidth}{@{\extracolsep{\fill}}lccccccc@{}}
\toprule
& \multicolumn{4}{c}{Primary unit} & Scope & \multicolumn{2}{c}{Verification} \\
\cmidrule(lr){2-5}\cmidrule(lr){6-6}\cmidrule(l){7-8}
Benchmark & App & WS & Ops & Fn & 1-Fn & Build & State \\
\midrule
\multicolumn{8}{@{}l}{\itshape Office applications and workspaces} \\
\cmidrule(lr){1-8}
\quad OfficeBench
  & \umark & \uoff & \uoff & \uoff & \nmark & \nmark & \cmark \\
\quad OdysseyBench
  & \umark & \uoff & \uoff & \uoff & \nmark & \nmark & \pmark \\
\quad Workspace-Bench
  & \uoff & \umark & \uoff & \uoff & \nmark & \nmark & \nmark \\
\midrule
\multicolumn{8}{@{}l}{\itshape Enterprise records and workflows} \\
\cmidrule(lr){1-8}
\quad WorkBench/Revisited
  & \uoff & \uoff & \umark & \uoff & \nmark & \nmark & \cmark \\
\quad CRMArena-Pro
  & \uoff & \uoff & \umark & \uoff & \pmark & \nmark & \nmark \\
\quad EnterpriseBench
  & \uoff & \uoff & \uoff & \umark & \pmark & \nmark & \pmark \\
\quad EntWorld
  & \umark & \uoff & \uoff & \uoff & \nmark & \cmark & \cmark \\
\quad Anchor/ERP-Bench
  & \uoff & \uoff & \umark & \uoff & \pmark & \cmark & \cmark \\
\midrule
\textbf{OfficePrism}
  & \uoff & \uoff & \uoff & \umark & \cmark & \cmark & \cmark \\
\bottomrule
\end{tabular*}
\end{table}
\endgroup

\noindent\textit{Stateful enterprise operations and diagnostic structure.}
AppWorld and $\tau$-bench use state checks, while WorkBench and WorkBench
Revisited compare final database state against ground truth
~\citep{trivedi2024appworld,yao2024taubench,styles2024workbench,
styles2026workbenchrevisited}. Enterprise benchmarks add linked records,
access hierarchies, workflow effects, and SQL, state-difference, or end-state
checks~\citep{huang2025crmarena,huang2026crmarena,
vishwakarma2025enterprisebench,gupta2026worldworkflows,
malay2026enterpriseopsgym,pysklo2026agentdiff}, while AutomationBench studies
search-then-execute tool discovery~\citep{shepard2026automationbench}.
OfficePrism groups state-checked tasks by business function and separates
construction replay from post-run database verification. Holding function
fixed, L1--L3 vary tool-chain complexity, information dependencies, and
workflow structure.

\noindent\textit{Task construction and evaluation.}
Schema-grounded construction is established. EnterpriseBench and CRMArena ground tasks in organizational context and records~\citep{vishwakarma2025enterprisebench,huang2025crmarena}. EntWorld derives workflows and SQL checks from schemas~\citep{mo2026entworld}. EnterpriseLab replays and deduplicates trajectories~\citep{agarwal2026enterpriselab}. Anchor compiles certified solutions and state verifiers from constraints~\citep{ivanov2026anchor}. OfficePrism co-generates a ticket, candidate tool sequence, and expected response from task-local state and rules, then replays the sequence before quality filtering.

Automated evaluation introduces a separate evidence problem. Studies of
trajectory rewards and language-model judges find sensitivity to context,
rubrics, model choice, and adversarial inputs
~\citep{lu2025agentrewardbench,xu2025contextualjudgebench,
zheng2023judging,raina2024judgerobust}. Benchmark methodology likewise
distinguishes construct validity from artifact coverage and catalogs recurring
benchmark failure modes~\citep{bean2025constructvalidity,
mcgregor2025benchrisk}. OfficePrism therefore treats construction replay as
executability evidence, keeps state checks distinct, and scopes language-model
judgments to explicit criteria.

%% file: sections/benchmark.tex
\section{Benchmark Design}
\label{sec:benchmark}

OfficePrism operationalizes function-closed office operation by measuring whether an agent can complete a task within one business function and how performance varies across benchmark-defined levels of within-function task structure. We use \emph{business function} for an organizational responsibility and \emph{domain} for the corresponding state-and-tool module. Each task is assigned to exactly one of seven functions---administration, finance, human resources, information technology, legal affairs, procurement, or project work---and uses that domain together with a shared organizational base. The base is infrastructure rather than an eighth task domain.

Every task uses the shared base and exactly one domain environment, with no
records or tools from another business function. OfficePrism organizes these
tasks by function and by L1--L3 task structure, defined through reference-chain
complexity, information dependencies, and workflow structure.

\begin{figure*}[t]
\centering
\includegraphics[width=\textwidth]{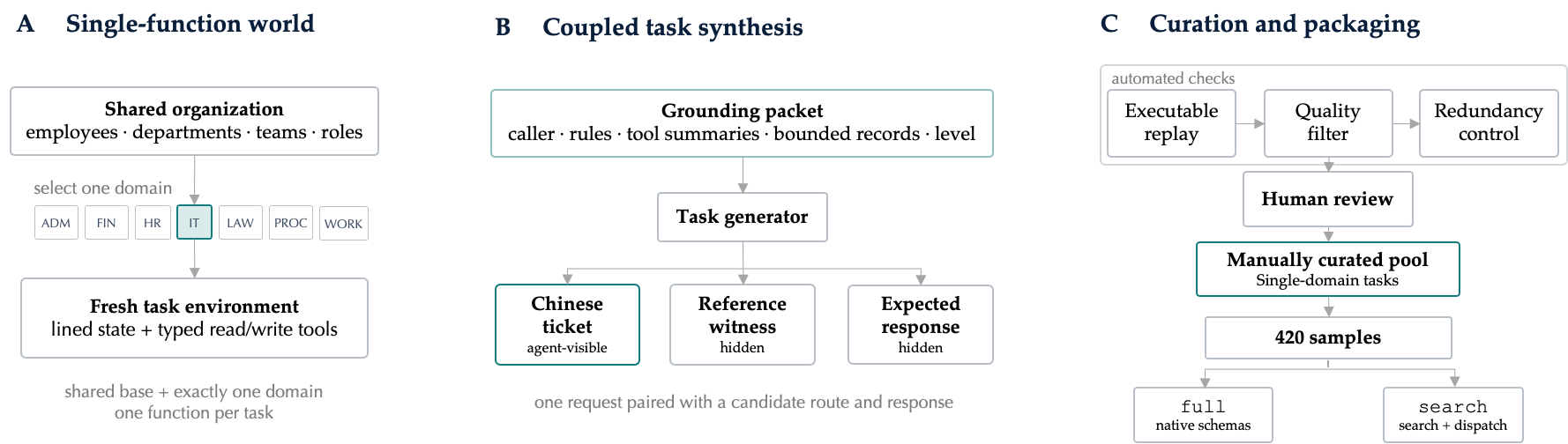}
\Description{A three-panel method overview. Panel A combines a shared
organization with exactly one of seven business domains to create a fresh
state and typed tool surface. Panel B builds a grounding packet from caller
context, rules, tool summaries, selected records, and a task-structure level, then
proposes a Chinese ticket, reference witness, and expected response. Panel C
applies executable replay, a quality filter, redundancy control, and human
review to form a manually curated pool before selecting
OfficePrism-420 and presenting it through full-schema or
search-and-dispatch interfaces.}
\caption{OfficePrism separates the single-function environment, coupled task synthesis, curation, benchmark selection, and the two evaluation interfaces.}
\label{fig:pipeline}
\end{figure*}

Figure~\ref{fig:pipeline} locates the visible and hidden task fields, replay,
filtering, curation, benchmark selection, and interface packages within the
end-to-end benchmark flow.

\subsection{Task Setting and Enterprise Environment}

Each evaluation run creates a fresh in-memory environment by loading a shared organizational base and one domain overlay. The base contains employees, departments, positions, teams, roles, and their relationships. A domain overlay contributes function-specific records, such as facilities and bookings for administration, accounts and reimbursements for finance, employee processes for human resources, system assets for information technology, contracts for legal affairs, suppliers and orders for procurement, or projects and work items for project work. Stable references connect each overlay to the same organization. The agent has no direct database access and can inspect or modify this state only through typed tools in the active environment.

Consider task \path{l2_323e0969} from OfficePrism-420. Its Chinese ticket asks the agent to identify the head of the design department, find that employee's primary email account, and enable a supplied out-of-office message. One reference witness resolves the department, identifies its head, lists that employee's email accounts, and updates the selected account. The later calls therefore depend on entities returned earlier, while the entire operation remains inside the shared organization and the information-technology domain. The ticket contains the business goal but not the internal references or an API recipe. Constructing a fresh environment for each trial prevents one trajectory from changing the initial state of the next.

\subsection{Task Record and Tool Interfaces}

The agent sees caller context and a Chinese natural-language ticket. Caller
context includes the available identity and organizational fields for the
requester; the ticket states the requested outcome.
Construction annotations remain hidden, including the fuller business
scenario, one reference witness, the expected response, and task metadata.
The agent acts through the exposed tools and may use any supported sequence.

OfficePrism provides \texttt{full} and \texttt{search} interface packages over
the same task-local toolkit. \texttt{full} exposes native schemas directly;
\texttt{search} uses discovery and dispatch to reach native operations. The
evaluation runner records the selected package and starts each task from an
isolated state. Section~\ref{sec:evaluation} describes their case-level
evaluation.

\subsection{Benchmark Scale and Composition}
\label{subsec:difficulty}

\begin{figure}[t]
  \centering
  \includegraphics[width=\columnwidth]{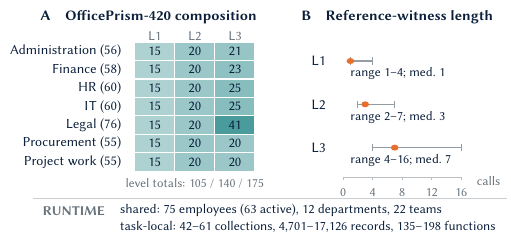}
  \caption{Composition and environment scale of OfficePrism-420.}
  \Description{The figure reports OfficePrism-420 task counts by business
  function and difficulty level, reference-witness length ranges and medians,
  and shared-organization and task-local environment scale.}
  \label{fig:task-profile}
\end{figure}

Figure~\ref{fig:task-profile} profiles a shared-plus-one-function runtime: a
75-employee organization, including 63 with active status, across 12
departments and 22 teams. Depending on the active function, the task-local
environment contains 42--61 top-level collections, 4,701--17,126 records, and
135--198 native functions.

\textbf{L1} expresses one direct intent. \textbf{L2} adds a dependent chain whose later call consumes an earlier result. \textbf{L3} combines at least two elements such as filtering, cross-entity transfer, conditional action, post-write checking, recovery, or aggregation. These structural labels impose no fixed call counts, and their empirical ordering is examined rather than assumed.

OfficePrism-420 contains 105 L1, 140 L2, and 175 L3 tasks. Every function
contributes 15 L1 and 20 L2 tasks; retained L3 counts range from 20 to 41
because the final selection preserves qualifying tasks from the fixed
development base before adding quota-targeted candidates. Reference witnesses
span 1--4, 2--7, and 4--16 calls for L1--L3, with medians of 1, 3, and 7.
Each witness records one executable construction route; evaluated agents may
use any supported route.

%% file: sections/construction.tex
\section{Benchmark Construction}
\label{sec:construction}

Constructing a useful task requires three properties at once. The ticket should read like an office request, the requested operation should be executable in a concrete enterprise state, and the workflow should expose the structure associated with its requested level. OfficePrism addresses these requirements by constraining generation with tools and records before writing the ticket, then retaining a serialized reference witness for construction-time replay. As summarized in Figure~\ref{fig:pipeline}, the pipeline consists of scenario organization, state-grounded task synthesis, automated checks, redundancy control, and human curation.

\subsection{Scenario-Conditioned Grounding}

Task construction begins from the tool and data inventory of one business domain. A configurable LLM survey receives tool names, docstrings, and table names, then proposes named tool--table groups around workplace objectives. The resulting inventory can be cached and reused. A parse failure falls back to one full-domain group. Groups containing only general organizational lookup functions are removed, and the remaining groups inherit the shared lookup tools needed to resolve employees, teams, and departments. L1 and L2 generation uses one scenario group. L3 generation samples and merges up to three groups from the same domain. These groups organize generation inputs rather than define a stable taxonomy of workplace activity.

For each requested task, the pipeline selects an active employee as caller and
builds a grounding packet. The packet contains caller context, domain rules,
the requested task-structure level, selected tool descriptions consisting of
names, parameter names, and truncated docstrings, six core organizational
tables, and the domain tables named by the scenario group. It serializes at
most the first ten stored records from each included table. All supplied
records and callable operations come from one shared-plus-domain environment.
The task record retains the selected tool and table names.

\subsection{Coupled Task Synthesis}

Writing a natural ticket first and attaching its construction annotations
afterward can produce drift: the request may appear plausible while its witness
or expected response is not supported by the same enterprise state.
OfficePrism instead proposes these artifacts together from one grounding
packet. The generator receives level-specific instructions and, when available,
at most one level-matched example selected against the current tool subset.
Selection prefers same-domain examples whose annotated tools are fully covered,
then those with the greatest positive tool overlap. If neither is available,
the implementation may use a fully covered, level-matched example from another
business function. This demonstration contributes no tools or records to the
generated task. The example guides workplace tone, context organization, and
implicit dependencies. The prompt forbids copying its entities, identifiers,
tool chain, numbers, or answer.

In one call, the prompt requests a business scenario, an agent-facing ticket
$q$, a reference witness $w$, and an expected operational response $y^\star$.
The ticket states the caller's goal without exposing tool or table names,
internal identifiers, or solution steps. The witness serializes one candidate
sequence of functions and arguments. The expected response states what should
be communicated after execution. All three fields are generated from the same
caller, rules, tools, and records.

Malformed outputs are discarded. Parsed candidates enter replay first, and
replay survivors enter quality filtering. A failed requested task slot may be
regenerated within configured retry and refill budgets.

\subsection{Automated Checks and Redundancy Control}

The first check replays the reference witness. A parser converts the serialized
chain into calls, verifies that every named tool exists, and executes the calls
sequentially on a deep copy of the initial environment. Empty or malformed
chains, unknown tools, invalid arguments, and runtime exceptions cause
rejection.

A second call to the configured model sees the requested level, caller name,
scenario, ticket, and witness. It scores logical coherence and absence of
answer shortcuts, workplace naturalness and absence of implementation leakage,
and alignment with the level-specific rubric. Each score must reach four on a
five-point scale.

The pipeline then applies two redundancy heuristics within a deduplication
batch. For candidates with the same set of reference-tool names, caller-role
bucket, and level, it retains the instance with the highest mean filter score.
It next compares normalized ticket text within caller-role and level buckets
using a configurable encoder. When no embedding model is supplied, the
implementation uses character-trigram cosine similarity with a threshold of
$0.85$.

\subsection{Human Curation and Pool Formation}

Human curation reviews the ticket, witness, and expected response as one
candidate case. Curators accept the case as written, repair its expected
response when the available evidence supports the requested outcome, or remove
it when no reproducible response can be recovered.

The supplied curation record reports a reduction from 852 candidates to a
manually curated pool of 574 single-domain tasks.

OfficePrism-420 is selected from this pool in a downstream step.
Section~\ref{sec:benchmark} reports its composition, and
Appendix~\ref{app:selection-provenance} reports selection and task-record
statistics.

%% file: sections/evaluation.tex
\section{Evaluation Protocol}
\label{sec:evaluation}

OfficePrism evaluates every system on the same 420 task records. Evaluation is
performed per case and only then aggregated: the runner iterates over task IDs,
creates a fresh domain environment for each trial, executes the agent, invokes
the evaluators on that task and trajectory, and stores a result row containing
the task ID, level, trial, component scores, termination reason, interaction
length, final response, interface mode, and trajectory. Thus, the percentages
reported in Section~\ref{sec:experiments} are aggregates of case-level
outcomes.

\subsection{Task Loading and Inputs}

Each record contains a Chinese request, an expected answer, and a reference
tool chain. The loader uses the expected answer for natural-language
evaluation and parses the tool chain into task-specific reference actions.

The agent receives the caller context, task request, domain policy, and the
tools exposed by the selected interface package. The expected answer, reference
calls, construction metadata, and evaluator criteria are not included in the
agent-facing ticket. Each task is evaluated in the environment for its single
business function, initialized from the same shared synthetic organization and
a function-specific state overlay. A new environment instance is constructed
for every task trial, preventing state written by one case from carrying into
the next.

\emph{Execution and record keeping.}
The same task loop supports both native and external agent harnesses. For each
case, it records the exposed and underlying tool counts, executes until the
harness terminates or reaches its step/error limit, and then evaluates the
resulting simulation. A normally completed record stores the component
breakdown alongside the trajectory rather than retaining only one scalar. If
agent execution raises an exception, the runner still emits a
task-ID-bearing failure record instead of silently dropping the case. Result
files are written after each case, so resumption can skip completed
task--trial pairs without changing the denominator of a completed 420-case
evaluation.

\subsection{Case-Level Scoring}

\emph{Natural-language outcome.}
The natural-language evaluator receives the recorded interaction and the
task-specific expected outcome. An LLM judge determines whether the trajectory
supports that outcome; if a task contains multiple assertions, their binary
decisions are aligned to the requested assertions and averaged, with an omitted
judge decision counted as a failure. The reported NL status is therefore an
expected-outcome judgment over the complete recorded interaction, including
tool observations and the final response.
The judge is required to return one JSON decision for each expected outcome.
Malformed output is treated as an empty result, and the alignment step inserts
an explicit failure for any missing outcome. This fail-closed behavior prevents
a partial judge response from receiving full credit.

\emph{Reference actions and state.}
The action evaluator extracts structured calls from the trajectory and matches
them against the task's parsed reference calls. For the \texttt{search}
package, a dispatcher call is expanded to its delegated native call before
matching. Reference-call coverage records whether the reference capabilities
appear in the trajectory without imposing a globally fixed order.
The matcher compares the native function name and the arguments named by the
reference action. A predicted call can satisfy at most one reference action,
but reference order is not scored. Action coverage therefore asks whether the
reference capabilities appeared in the trace, whereas the database component
asks whether the requested state was reached.

Database evaluation compares the state reconstructed from the agent trajectory
with the state obtained by replaying the task's reference calls. The evaluator
replays each sequence in a fresh environment and compares the resulting
database state. For the 104 tasks whose reference calls write Chinese free
text, DB status combines (i) structural state equality after normalizing 24
named free-text fields and (ii) a
\texttt{qwen3-max}, temperature-zero judgment of the written text. The text
check rejects factual fabrication, intent mismatch, an unidentifiable core
object, or an omitted value, while allowing wording and detail to differ. This
correction addresses both false negatives from harmless wording differences
and false positives from writes that did not take effect.

NL checks whether the interaction supports the requested response; DB compares
the final environment with reference replay; and action coverage records the
reference capabilities found in the trajectory. The headline combines NL and
DB. Stored
Success, Strict Success, reward, and step fields remain secondary diagnostics
reported in the appendix.

The headline outcome is defined for each task \(i\) and displayed configuration
\(c\) as
\[
  y_{ic}=\mathbb{1}\{\mathrm{NL}_{ic}=1
  \ \land\ \mathrm{DB}_{ic}=1\}.
\]
The OfficePrism-420 pass rate is the unweighted case mean
\[
  \mathrm{NL\mbox{-}DB}_{c}
  =\frac{1}{420}\sum_{i=1}^{420}y_{ic}.
\]
Level-specific rates apply the same definition to the fixed L1, L2, and L3
subsets of 105, 140, and 175 tasks. Read-only cases leave the database
unchanged, so their DB component checks preservation of the initial state;
write cases additionally require the expected structural and textual updates.

\subsection{Interface Packages and Aggregation}

The \texttt{full} package exposes the native schemas available in the active
business function. The \texttt{search} package presents the same underlying
function toolkit through discovery and dispatch. Both packages use the same
420 tasks and task-local environments. They differ in schema visibility,
discovery, dispatch, observations, errors, and interaction length.

The current result panel contains one 420-case aggregate for each of 124
reported model--harness--package configurations. We report the exact 420-case
rates together with medians, ranges, and sign counts across configurations.
Each configuration was evaluated once. Step fields report the interaction
length recorded by each harness and interface package.

%% file: sections/experiments.tex
\section{Experiments and Analysis}
\label{sec:experiments}

We report results on one benchmark artifact: OfficePrism-420. The analysis
first summarizes the complete 124-row display panel, then examines
benchmark-defined difficulty levels and the two interface packages. Every row
aggregates the case-level NL--DB outcomes of the same 420 tasks.

\subsection{Experimental Setup}
\label{subsec:experimental-setup}

\emph{Benchmark.}
The benchmark contains 420 tasks across seven business functions, with each
task confined to one domain. Its level composition is 105 L1, 140 L2, and
175 L3 tasks. The same checksummed 420-task JSON is used throughout this
section. The set retains all 300 records from the earlier collection and adds
120 records. The additions were selected using pilot performance, with a tie
preference for search-friendly cases.

\emph{Systems and interfaces.}
The leaderboard contains 124 aggregate rows spanning five harness
display labels (ReAct, DeepAgents, Claude Code, OpenClaw, and Hermes), 14 model
display labels, and two complete interface packages. The \texttt{full}
package exposes the complete tool schema, whereas \texttt{search} presents
the search-and-dispatch interface defined in Section~\ref{sec:evaluation}.
There are 54 \texttt{full} rows and 70 \texttt{search} rows. Every
harness--package cell contains the same ten-model core; \texttt{search}
contains four additional models for every harness, while DeepAgents also
contains those four under \texttt{full}. Display labels identify the evaluated
configurations reported in the summary.

\emph{Outcomes.}
We take NL--DB Pass as the headline case-level outcome. For 104 tasks whose
reference calls contain Chinese free text, database status combines structural
replay---with 24 named free-text fields normalized for comparison---and a
\texttt{qwen3-max} text judgment at temperature zero. That judgment penalizes
factual error or fabrication, intent mismatch, an unidentifiable core object,
and omitted text, but not differences in wording or detail.

We preserve the reported L1--L3 and average-step fields as diagnostics and
disclose Success and Strict Success only in the appendix. Across all 124 rows,
Success equals 100 times average NL and Strict Success equals 100 times average
reward at the supplied precision. Weighting the level rates by the fixed
105/140/175 task counts reconstructs NL--DB Pass up to one-decimal rounding,
confirming that the table uses the advertised 420-case denominator.

Each displayed configuration is represented by one evaluation of the 420-case
set. We report exact task-set rates and medians, ranges, and sign counts across
rows. Appendix~\ref{app:displayed-results}
reproduces all 124 rows, and the accompanying audit records the exact input
hashes and every transformation used to generate the tables and figures.

\subsection{Main Results}
\label{subsec:main-results}

\begin{figure*}[t]
  \centering
  \includegraphics[width=\textwidth]{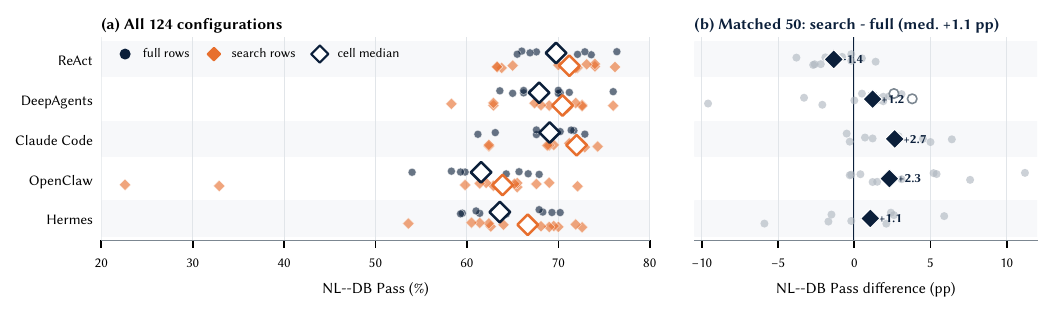}
  \caption{OfficePrism-420 result panel. (a) All 124 NL--DB Pass
  rows, with vertically offset \texttt{full} and \texttt{search} bands and
  open diamonds for unweighted cell medians. (b) Search-minus-full
  differences; filled points are the common-support 50-configuration panel,
  open points are four additional DeepAgents comparisons, and diamonds mark
  common-support medians.}
  \Description{Two aligned horizontal dot plots. The left panel shows all
  124 NL--DB Pass rows by harness, with full and search rows
  separated by color, shape, and vertical offset. The right panel shows
  matched-label score differences on the same five harness rows; points occur on
  both sides of zero and harness medians differ in sign.}
  \label{fig:e6-main-results}
\end{figure*}

Across the 124 display rows, NL--DB Pass ranges from 22.6\% to 76.4\%, with
an unweighted median of 67.9\% and mean of 66.8\%.
Figure~\ref{fig:e6-main-results}(a) shows every row on a common horizontal
scale. Open diamonds mark within-cell medians; the vertically offset package
bands show the two interface packages.

The upper end is shared by more than one harness and package. The best
displayed configuration is ReAct with \texttt{kimi-k3} under \texttt{full} at
76.4\%; the same model reaches 76.2\% with ReAct under \texttt{search} and
76.0\% with DeepAgents under either package. The next group contains Claude
Code with \texttt{glm-5.2} under \texttt{search} at 74.3\%, followed by two
ReAct--search rows at 74.0\%. Thus, the top four differ by only 0.4 points,
whereas changing the surrounding harness can produce much larger differences
for other model labels.

The lower tail is similarly configuration-specific. Two OpenClaw
\texttt{search} rows score 32.9\% and 22.6\%, while every other row is at least
53.6\%. These two entries produce the wide lower tail visible in
Figure~\ref{fig:e6-main-results}; the complete row-level display shows both
this tail and the central mass.

The first and third quartiles of the 124-row panel are 63.3\% and 71.2\%.
In total, 113 rows reach at least 60\%, 86 reach at least 65\%, 46 reach at
least 70\%, 23 reach at least 72\%, and seven reach at least 74\%. The seven
rows above the last threshold come from ReAct, DeepAgents, and Claude Code
configurations under the two packages.

\input{generated/e6_main_results_table}

Table~\ref{tab:e6-main-results} summarizes the same panel by
harness--package cell. Each value is the unweighted median of the displayed
rows in that cell; the All column is not an average of the three level
medians. The row counts in Table~\ref{tab:e6-main-results} show the unequal
model support across package cells.

\subsection{Performance Across Task-Structure Levels}
\label{subsec:difficulty-results}

Across all 124 rows, the unweighted median NL--DB Pass is 83.8\% for L1,
71.8\% for L2, and 56.6\% for L3. L1 exceeds both L2 and L3 in every row;
L2 exceeds L3 in 122 of 124 rows. The appendix tables disclose every
row-level L1--L3 value; we omit a second plot of the same entries.

The magnitude of the separation is not confined to a few configurations. The
within-row L1--L2 gap has a median of 11.9 percentage points and ranges from
2.7 to 23.1; the L2--L3 gap has a median of 16.3 and ranges from \(-5.3\) to
35.0. Consequently, the median L1--L3 gap is 27.5 points, with a range of 9.9
to 46.9. The two negative L2--L3 differences are the same two exceptions to
the 122-of-124 ordering reported above.

The task records differ along several measured dimensions. Every function
contributes 15 L1 and 20 L2 tasks, while its L3 count ranges from 20 to 41.
Mean reference-witness length increases from 1.50 calls at L1 to 3.34 at L2
and 7.66 at L3; mean question length increases from 39 to 92 and 209 Chinese
characters; and mean expected-response length is 130, 147, and 390
characters. Thus, the level-wise score gaps accompany longer requests,
responses, and reference sequences.

\subsection{Outcome Components and Interaction Length}
\label{subsec:component-results}

The result rows provide the natural-language and database components alongside
the joint outcome. Across all 124 configurations, the natural-language rate
ranges from 24.8\% to 85.5\%, with a median of 74.9\%. The database rate ranges
from 73.6\% to 94.5\%, with a median of 82.6\%. The corresponding joint
NL--DB median is 67.9\%, and Strict Success ranges from 18.3\% to 61.4\%,
with a median of 53.8\%.

\input{generated/e6_outcome_components_table}

Table~\ref{tab:e6-outcome-components} separates these quantities for every
harness--package cell. The cell medians span 68.3--78.8\% for NL,
80.3--85.2\% for DB, 61.5--72.0\% for NL--DB, and 49.0--57.6\% for Strict
Success. The table also shows that median recorded interaction length ranges
from 7.3 to 19.5 steps across the ten cells.

\begin{figure}[htb]
  \centering
  \includegraphics[width=\columnwidth]{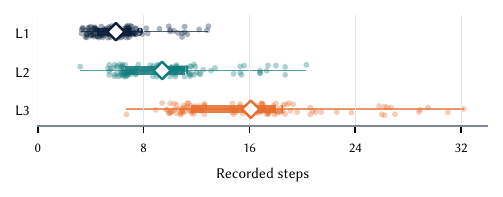}
  \caption{Recorded interaction steps by task-structure level across the 124
  configurations. Points are configurations, thick bars span the
  interquartile range, diamonds mark medians, and thin lines show the observed
  range.}
  \Description{Three compact horizontal dot distributions show recorded
  interaction steps for L1, L2, and L3. Their medians increase from 5.9 to
  9.4 and 16.1 steps.}
  \label{fig:e6-level-steps}
\end{figure}

Interaction length also varies with the task-structure level.
Figure~\ref{fig:e6-level-steps} shows the distribution of the supplied
level-specific step averages across configurations. The median rises from 5.9
steps at L1 to 9.4 at L2 and 16.1 at L3. Within individual configurations, the
L2 average exceeds L1 in 123 of 124 rows, and L3 exceeds L2 in all 124 rows.
The median within-row increases are 3.4 steps from L1 to L2 and 6.3 from L2
to L3.

\subsection{Variation Across Agent Harnesses}
\label{subsec:harness-results}

The panel also permits a common-support check of how strongly the harness
label conditions a model result. Ten model labels occur under both packages
for all five harnesses, producing 20 model--package groups with five harness
scores each. Within a group, the maximum-minus-minimum NL--DB range has a
median of 8.6 percentage points; all 20 ranges exceed 1.1 points, and 17 exceed
5 points. The smallest range is 3.5 points for \texttt{kimi-k2.6} under
\texttt{search}. The largest is 15.4 points for \texttt{qwen3.6-plus} under
\texttt{search}, whose displayed scores span 53.6--69.0\%. Under
\texttt{full}, \texttt{qwen3.6-max} spans 54.0--68.6\%, a 14.6-point range.

These ranges are substantially larger than the \(+1.1\)-point median package
difference reported below. The same model--package label therefore records
different scores across harnesses in this panel. We retain
model--harness--package as the reporting unit.

\subsection{Common-Support Model Profiles}
\label{subsec:common-model-results}

Ten model labels appear under both packages for every harness. Each of these
labels therefore contributes ten directly comparable display rows: one for
each of five harnesses under \texttt{full} and \texttt{search}.
Table~\ref{tab:e6-common-models} summarizes the resulting score distributions
without adding the four labels that appear in only six harness--package rows.

\input{generated/e6_common_model_table}

\texttt{kimi-k3} has the highest ten-row median at 72.5\%, followed by
\texttt{glm-5.2} at 71.1\%. Their displayed ranges are 66.7--76.4\% and
67.6--74.3\%, respectively. Across the remaining labels, medians range from
61.9\% to 68.7\%. The narrowest maximum--minimum span is 6.7 points for
\texttt{glm-5.2}; the widest are 17.2 points for \texttt{qwen3.6-max} and
15.9 points for \texttt{qwen3.6-plus}.

The last two columns separately aggregate the five harness rows under each
package. The \texttt{search} median exceeds the \texttt{full} median for nine
of the ten labels; \texttt{qwen3.6-plus} is the exception, with medians of
63.8\% and 67.6\%. These per-label medians complement the paired row-level
comparison below, which retains all 50 individual harness--model pairs rather
than first aggregating over harnesses.

\subsection{Full/Search Package Comparison}
\label{subsec:interface-results}

All 54 \texttt{full} rows have a matching \texttt{search} display label.
Across these comparisons, search is higher in 34 cases, tied in one, and
lower in 19; the search-minus-full difference has a median of \(+1.2\) and a
mean of \(+1.13\) percentage points. DeepAgents contributes 14 comparisons,
while each other harness contributes 10.

For equal support across harnesses, we use the ten model labels available
under both packages for every harness, yielding 50 comparisons
(Figure~\ref{fig:e6-main-results}(b)). Search is higher in 30, tied in one,
and lower in 19, with a median difference of \(+1.1\) and mean difference of
\(+1.04\) points. The harness-specific medians vary in sign: \(-1.4\) points
for ReAct, \(+1.2\) for DeepAgents, \(+2.7\) for Claude Code, \(+2.3\) for
OpenClaw, and \(+1.1\) for Hermes.

The dispersion is large relative to the \(+1.1\)-point pooled median. Across
the 50 comparisons, differences range from \(-9.6\) to \(+11.2\)
points, and every harness contains both positive and negative differences.
The positive/tied/negative counts are \(2/0/8\) for ReAct, \(6/1/3\) for
DeepAgents, \(8/0/2\) for Claude Code, \(8/0/2\) for OpenClaw, and \(6/0/4\)
for Hermes. The pooled direction therefore hides substantial model-level
variation even after restricting the display to common labels.

The marginal cell summaries in Table~\ref{tab:e6-main-results} reinforce why
the estimand matters. The displayed \texttt{search} median is higher in every
harness cell, yet the balanced paired median is negative for ReAct. These
statements are compatible because the cell medians summarize package-specific
row sets, whereas the paired diagnostic conditions on the ten shared model
labels. In particular, four additional \texttt{search} labels enter each
marginal cell, while only DeepAgents also reports those labels under
\texttt{full}.

We also compare the ordering of the ten common model labels within each
harness. Table~\ref{tab:e6-package-rank} reports Pearson correlation for the
displayed scores and Spearman correlation for their ranks under
\texttt{full} and \texttt{search}. All five correlations are positive.
Spearman correlation ranges from \(0.637\) for Claude Code to \(0.912\) for
ReAct; Pearson correlation ranges from \(0.544\) for OpenClaw to \(0.946\)
for ReAct.

\input{generated/e6_package_rank_table}

The packages also differ in recorded interaction length. \texttt{Search}
reports more average steps in all 50 comparisons: the search-minus-full
difference ranges from \(+3.3\) to \(+8.6\) steps, with a median of \(+4.7\).
The harness-specific medians are \(+7.5\) for ReAct, \(+4.3\) for DeepAgents,
\(+4.0\) for Claude Code, \(+5.4\) for OpenClaw, and \(+4.6\) for Hermes.

\input{generated/e6_matched_step_table}

Table~\ref{tab:e6-matched-steps} separates the same matched comparison by
task-structure level. Search records more steps in every one of the 50 pairs
at L1, L2, and L3. The median difference grows from \(+2.6\) steps at L1 to
\(+4.3\) at L2 and \(+6.5\) at L3. The largest observed difference is
\(+5.0\) at L1, \(+8.0\) at L2, and \(+11.6\) at L3.

%% file: generated/e6_main_results_table.tex
\begin{table}[htb]
\caption{Median results by harness and interface package.}
\label{tab:e6-main-results}
\centering
\footnotesize
\setlength{\tabcolsep}{2.2pt}
\renewcommand{\arraystretch}{1.06}
\begin{tabular*}{\columnwidth}{@{\extracolsep{\fill}}llrrrrrr@{}}
\toprule
Harness & Pkg. & Rows & L1 & L2 & L3 & All & Steps \\
\midrule
ReAct & \texttt{full} & 10 & 85.8 & 75.3 & 59.5 & 69.8 & 12.3 \\
ReAct & \texttt{search} & 14 & 85.7 & 74.7 & 57.4 & 71.2 & 19.5 \\
\addlinespace[1.2pt]
DeepAgents & \texttt{full} & 14 & 82.8 & 71.8 & 56.6 & 67.9 & 7.3 \\
DeepAgents & \texttt{search} & 14 & 85.7 & 74.3 & 57.4 & 70.5 & 11.8 \\
\addlinespace[1.2pt]
Claude Code & \texttt{full} & 10 & 85.8 & 72.8 & 56.5 & 69.0 & 8.0 \\
Claude Code & \texttt{search} & 14 & 87.2 & 76.1 & 58.0 & 72.0 & 11.9 \\
\addlinespace[1.2pt]
OpenClaw & \texttt{full} & 10 & 75.7 & 64.7 & 52.6 & 61.5 & 8.1 \\
OpenClaw & \texttt{search} & 14 & 81.5 & 68.6 & 50.3 & 63.9 & 12.8 \\
\addlinespace[1.2pt]
Hermes & \texttt{full} & 10 & 78.5 & 66.0 & 56.9 & 63.6 & 8.0 \\
Hermes & \texttt{search} & 14 & 81.0 & 70.7 & 56.0 & 66.7 & 12.6 \\
\bottomrule
\end{tabular*}
\end{table}

%% file: generated/e6_outcome_components_table.tex
\begin{table}[htb]
\caption{Median outcome components by harness and interface package. NL, DB,
NL--DB, and Strict are percentages; Steps is the recorded interaction length.}
\label{tab:e6-outcome-components}
\centering
\footnotesize
\setlength{\tabcolsep}{2.7pt}
\renewcommand{\arraystretch}{1.04}
\begin{tabular*}{\columnwidth}{@{\extracolsep{\fill}}llrrrrrr@{}}
\toprule
Harness & Pkg. & Rows & NL & DB & NL--DB & Strict & Steps \\
\midrule
ReAct & \texttt{full} & 10 & 78.8 & 83.9 & 69.8 & 55.3 & 12.3 \\
ReAct & \texttt{search} & 14 & 78.1 & 83.6 & 71.2 & 57.6 & 19.5 \\
\addlinespace[1.2pt]
DeepAgents & \texttt{full} & 14 & 75.3 & 80.3 & 67.9 & 53.6 & 7.3 \\
DeepAgents & \texttt{search} & 14 & 77.0 & 82.3 & 70.5 & 54.4 & 11.8 \\
\addlinespace[1.2pt]
Claude Code & \texttt{full} & 10 & 75.8 & 80.5 & 69.0 & 56.7 & 8.0 \\
Claude Code & \texttt{search} & 14 & 77.8 & 82.3 & 72.0 & 56.9 & 11.9 \\
\addlinespace[1.2pt]
OpenClaw & \texttt{full} & 10 & 68.3 & 81.5 & 61.5 & 49.1 & 8.1 \\
OpenClaw & \texttt{search} & 14 & 70.4 & 81.8 & 63.9 & 51.2 & 12.8 \\
\addlinespace[1.2pt]
Hermes & \texttt{full} & 10 & 70.6 & 84.3 & 63.6 & 49.0 & 8.0 \\
Hermes & \texttt{search} & 14 & 71.8 & 85.2 & 66.7 & 51.7 & 12.6 \\
\bottomrule
\end{tabular*}
\end{table}

%% file: generated/e6_common_model_table.tex
\begin{table}[htb]
\caption{NL--DB Pass across the ten harness--package configurations available
for each common-support model label.}
\label{tab:e6-common-models}
\centering
\footnotesize
\setlength{\tabcolsep}{2.2pt}
\renewcommand{\arraystretch}{1.04}
\begin{tabular*}{\columnwidth}{@{\extracolsep{\fill}}lrrrrr@{}}
\toprule
Model & Median & Min & Max & Full med. & Search med. \\
\midrule
\texttt{kimi-k3} & 72.5 & 66.7 & 76.4 & 70.2 & 72.9 \\
\texttt{glm-5.2} & 71.1 & 67.6 & 74.3 & 70.2 & 71.9 \\
\texttt{qwen3.7-plus} & 68.7 & 59.8 & 72.9 & 67.9 & 70.0 \\
\texttt{deepseek-v4-pro} & 68.1 & 65.5 & 72.9 & 67.6 & 68.1 \\
\texttt{glm-5.1} & 68.0 & 62.9 & 72.4 & 67.1 & 69.0 \\
\texttt{qwen3.7-max} & 66.5 & 61.4 & 72.6 & 65.5 & 69.0 \\
\texttt{deepseek-v4-flash} & 66.3 & 61.7 & 73.6 & 66.2 & 66.4 \\
\texttt{qwen3.6-max} & 66.1 & 54.0 & 71.2 & 63.6 & 66.9 \\
\texttt{qwen3.6-plus} & 64.2 & 53.6 & 69.5 & 67.6 & 63.8 \\
\texttt{kimi-k2.6} & 61.9 & 58.3 & 66.0 & 61.2 & 62.4 \\
\bottomrule
\end{tabular*}
\end{table}

%% file: generated/e6_package_rank_table.tex
\begin{table}[htb]
\caption{Agreement between \texttt{full} and \texttt{search} NL--DB scores
over the ten common model labels in each harness.}
\label{tab:e6-package-rank}
\centering
\footnotesize
\setlength{\tabcolsep}{5pt}
\begin{tabular*}{\columnwidth}{@{\extracolsep{\fill}}lrrr@{}}
\toprule
Harness & \(N\) & Pearson \(r\) & Spearman \(\rho\) \\
\midrule
ReAct & 10 & 0.946 & 0.912 \\
DeepAgents & 10 & 0.617 & 0.664 \\
Claude Code & 10 & 0.797 & 0.637 \\
OpenClaw & 10 & 0.544 & 0.644 \\
Hermes & 10 & 0.823 & 0.855 \\
\bottomrule
\end{tabular*}
\end{table}

%% file: generated/e6_matched_step_table.tex
\begin{table}[htb]
\caption{Search-minus-full recorded steps for the 50 common-support
harness--model pairs.}
\label{tab:e6-matched-steps}
\centering
\footnotesize
\setlength{\tabcolsep}{4.5pt}
\begin{tabular*}{\columnwidth}{@{\extracolsep{\fill}}lrrr@{}}
\toprule
Step field & Median & Mean & Range \\
\midrule
L1 & \(+2.6\) & \(+2.85\) & \(+1.5\) to \(+5.0\) \\
L2 & \(+4.3\) & \(+4.62\) & \(+2.1\) to \(+8.0\) \\
L3 & \(+6.5\) & \(+6.95\) & \(+2.6\) to \(+11.6\) \\
All & \(+4.7\) & \(+5.15\) & \(+3.3\) to \(+8.6\) \\
\bottomrule
\end{tabular*}
\end{table}

%% file: sections/limitations_conclusion.tex
\section{Limitations and Conclusion}
\subsection{Limitations and Responsible Use}
\label{sec:limitations}

OfficePrism studies function-closed operations in one synthetic organization,
with Chinese-language requests and a fixed, selected 420-task panel. It does
not estimate the prevalence of workplace activities or represent the
distribution of organizations, languages, roles, or business processes.
L1--L3 are benchmark-defined task-structure levels rather than a validated
causal or monotonic difficulty scale. The 124 reported configurations are
finite-set summaries with one aggregate record per configuration; they do not
provide repeated-run uncertainty. Likewise, the full/search comparison is a
descriptive contrast between interface packages, not an isolated causal
effect. Reference-witness replay shows that one retained call sequence
executes under implemented preconditions, but it does not prove that the route
is unique or optimal, nor does it by itself establish deployment reliability.

The benchmark is intended for research on agent evaluation and tool-mediated
reasoning in synthetic environments. Synthetic records reduce direct exposure
of real employee data, but generated attributes and organizational relations
can still encode unrealistic assumptions or sensitive proxies; the benchmark
therefore provides no privacy, fairness, or safety guarantee for real
workplaces. It should not be used for hiring, employee ranking or monitoring,
disciplinary decisions, access-control decisions, or unsupervised execution in
production systems. Any operational adaptation requires explicit
authorization, least-privilege credentials, security and privacy review,
domain-expert validation, independent fairness assessment, auditable logs, and
human control over consequential actions.

\subsection{Conclusion}
\label{sec:conclusion}

OfficePrism provides 420 function-closed tasks across seven business
functions with fresh task-local state, typed tools, and coupled tickets,
witnesses, and expected responses. Across 124 model--harness--package
configurations, NL--DB Pass ranges from 22.6\% to 76.4\%, with a median of
67.9\%; L1 is highest in all 124 configurations, and L2 exceeds L3 in 122.
Among 50 matched model--harness pairs, \texttt{search} scores higher in 30,
ties in one, scores lower in 19, and records more steps in all 50. The
evaluation uses one synthetic organization, Chinese requests, and one
420-case run per configuration. Upon acceptance, we will release the task
artifact, evaluator and runner code, interface documentation, and experiment
configurations.

%% file: sections/acknowledgement.tex
\begin{acks}
This work was supported by \textbf{Alibaba Group through Alibaba Innovative Research Program}. We would also like to express our sincere gratitude to \textbf{Mengni Ye}, \textbf{Yilun Zhang}, and \textbf{Peixin Lin}, technical experts at Alibaba Group, for their invaluable guidance, technical support, and insightful discussions throughout this work.
\end{acks}

%% file: sections/appendix.tex
\section{Extended Result Diagnostics}
\label{app:extended-diagnostics}

This appendix reports seven complementary views of agent performance on an extended 420-task display set, consisting of single-domain tasks drawn from the same seven business functions and L1--L3 levels. Unless stated otherwise, every value is a harness--model--package aggregate of the joint natural-language-and-database criterion (NL-DB Pass) on this set. Together, the panels substantiate the central claims of the paper: the benchmark retains substantial headroom for the strongest current systems, function-resolved scores expose structure that a single overall number hides, the two interface packages trade accuracy against interaction cost in measurable ways, a per-failure root-cause audit shows where the remaining errors concentrate, and two in-depth case studies trace how such failures escape conventional evaluation.

\subsection{Overall Leaderboard}
\label{app:ext-leaderboard}

\begin{figure*}[p]
  \centering
  \includegraphics[width=0.98\textwidth]{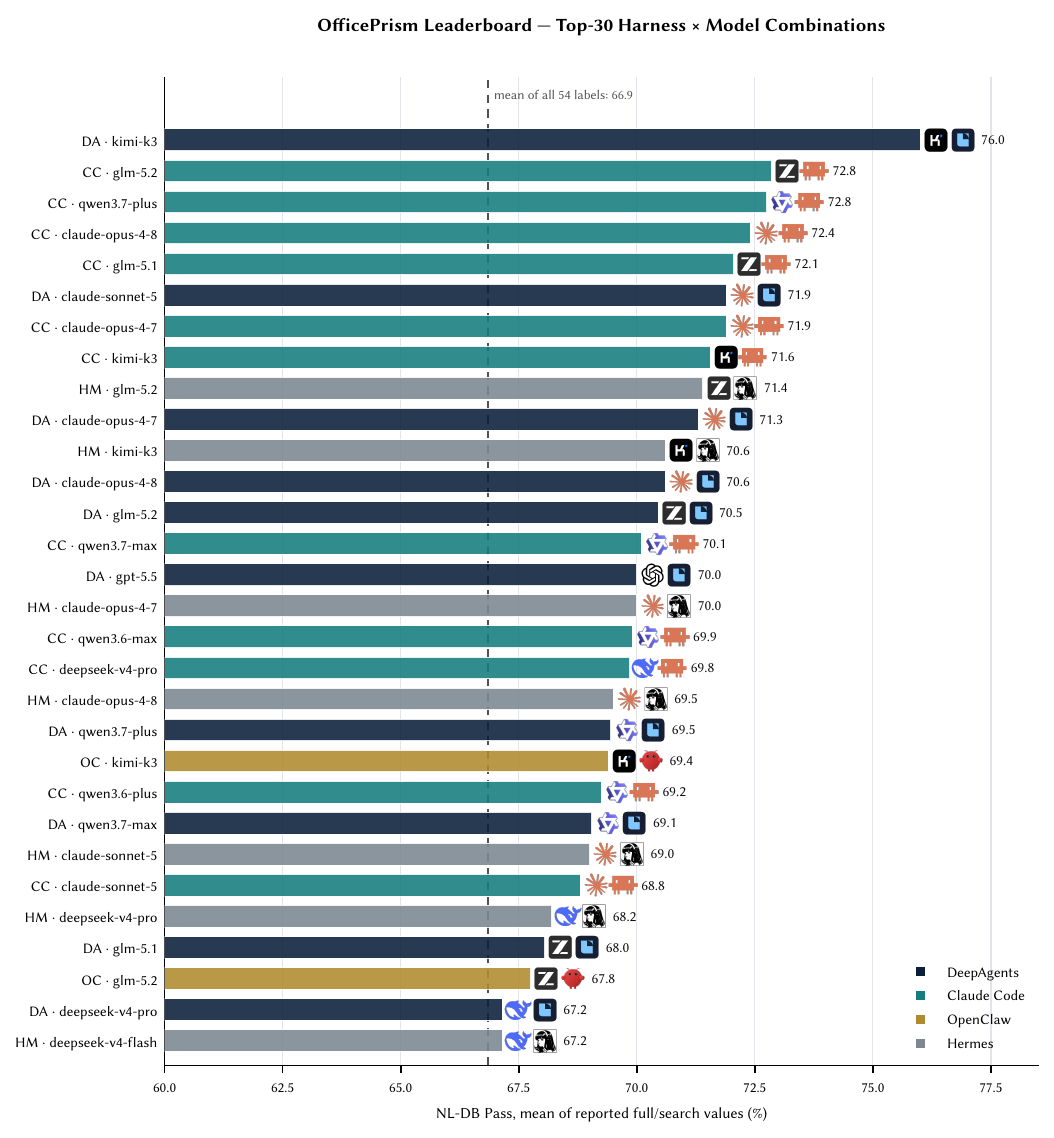}
  \caption{Top-30 harness--model combinations by mean full/search NL-DB Pass on the 420-task display set. Each bar end shows the model-family and harness marks followed by the one-decimal score, and the dashed line marks the mean of all 54 combinations. Harness abbreviations: DA (DeepAgents), CC (Claude Code), OC (OpenClaw), HM (Hermes); model names are shown in full.}
  \Description{A full-page horizontal bar chart ranks thirty harness and model combinations from highest to lowest mean NL-DB Pass. Bars are colored by harness, each bar end shows a model-family mark, a harness mark, and a one-decimal percentage, and a dashed vertical line marks the overall mean.}
  \label{fig:e6b-leaderboard}
\end{figure*}

Figure~\ref{fig:e6b-leaderboard} ranks the thirty strongest harness--model combinations by the mean of their full-package and search-package NL-DB values; combinations evaluated under a single package keep that value, and the ReAct is not ranked. Two observations support the benchmark's diagnostic purpose. First, headroom remains large: the best combination reaches 76.0\%, the mean over all 54 ranked combinations is only 66.9\% (dashed line), and no combination approaches the ceiling, so single-function operation is far from solved even before any cross-function coordination is demanded. Second, the harness contributes on a scale comparable to the model: the same Kimi~k3 model spans roughly seven points across harnesses (76.0 under DeepAgents versus 69.4 under OpenClaw), and the top of the ranking is dominated by the DeepAgents and Claude Code harnesses rather than by any single model family. Evaluations that report a model score without fixing the harness therefore conflate two distinct sources of capability.

\subsection{Performance by Business Domains}
\label{app:ext-function-radar}

\begin{figure*}[!t]
  \centering
  \includegraphics[width=\textwidth]{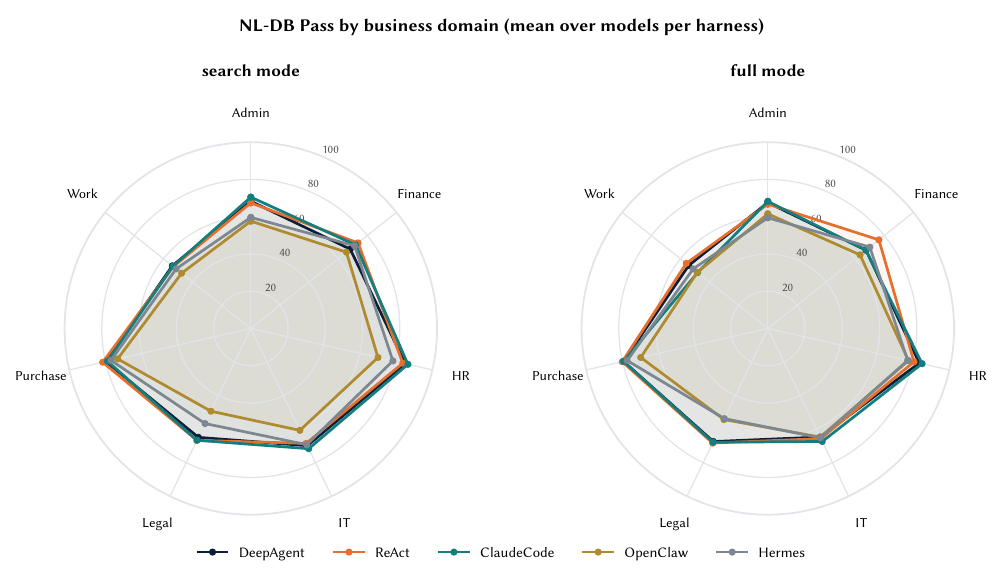}
  \caption{Level-reweighted NL-DB Pass by business function, averaged over models within each harness, for the search and full packages.}
  \Description{Two radar charts with seven business-function axes compare five harness profiles, the search package on the left and the full package on the right, on a zero to one hundred radial scale.}
  \label{fig:e6b-domain-radar}
\end{figure*}

Domain-resolved evaluation is the core design decision of OfficePrism, and Figure~\ref{fig:e6b-domain-radar} shows why it matters empirically. Each radar axis is one of the seven business domains, each profile averages the models within one harness. The spread across domains is far larger than the spread across systems: averaged over harnesses, human resources reaches 80.7\% while project work stays near 51.6\%, a gap of almost thirty points that a single overall score would flatten completely. The function ordering is essentially stable across the two interface packages and across harnesses, which indicates that the difficulty is a property of the domain's records and workflows rather than of any particular system, and it identifies project work and legal affairs as the natural targets for future capability work.

\subsection{Token--Performance Trade-off}
\label{app:ext-token-performance}

\begin{figure*}[!t]
  \centering
  \includegraphics[width=\textwidth]{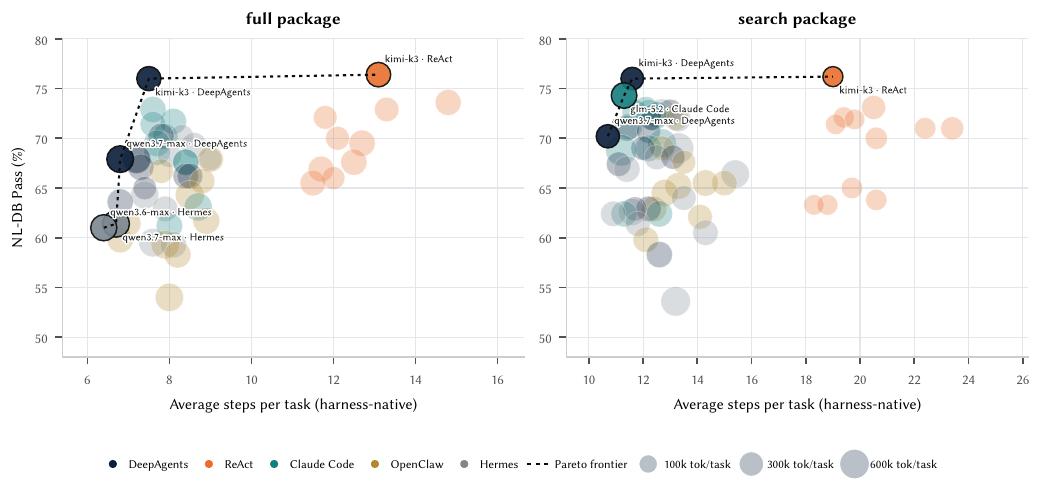}
  \caption{NL-DB Pass against average steps per task, with bubble area proportional to tokens per task, for the full and search packages. The dotted line connects the labeled Pareto-frontier combinations; dominated combinations are faded.}
  \Description{Two bubble panels plot NL-DB Pass against average steps for the full and search packages. Bubble color encodes harness and bubble area encodes tokens per task. In each panel a dotted line connects the labeled Pareto-frontier bubbles, and the other bubbles are drawn faded.}
  \label{fig:e6b-token-performance}
\end{figure*}

Accuracy alone does not determine whether an agent is deployable, so Figure~\ref{fig:e6b-token-performance} relates NL-DB Pass to interaction cost. Each bubble is one harness--model combination: the horizontal position is its average number of steps per task, the vertical position its NL-DB Pass, and the bubble area its average tokens per task. The dotted line connects the cost--performance Pareto frontier, the combinations for which no alternative is simultaneously faster and more accurate; frontier bubbles are labeled and dominated bubbles are faded. The frontier shows that the highest accuracy does not require the largest budget: Kimi~k3 under DeepAgents anchors the frontier in both packages, reaching the top accuracy band in roughly half the steps that ReAct-style prompting needs for comparable scores. Steps and tokens are counted by each harness in its own units, so cross-harness comparisons are indicative rather than exact, but the within-package frontier makes clear which combinations pay for their accuracy and which do not.

\subsection{Full versus Search per Combination}
\label{app:ext-interface-dumbbell}

\begin{figure*}[!t]
  \centering
  \includegraphics[width=\textwidth]{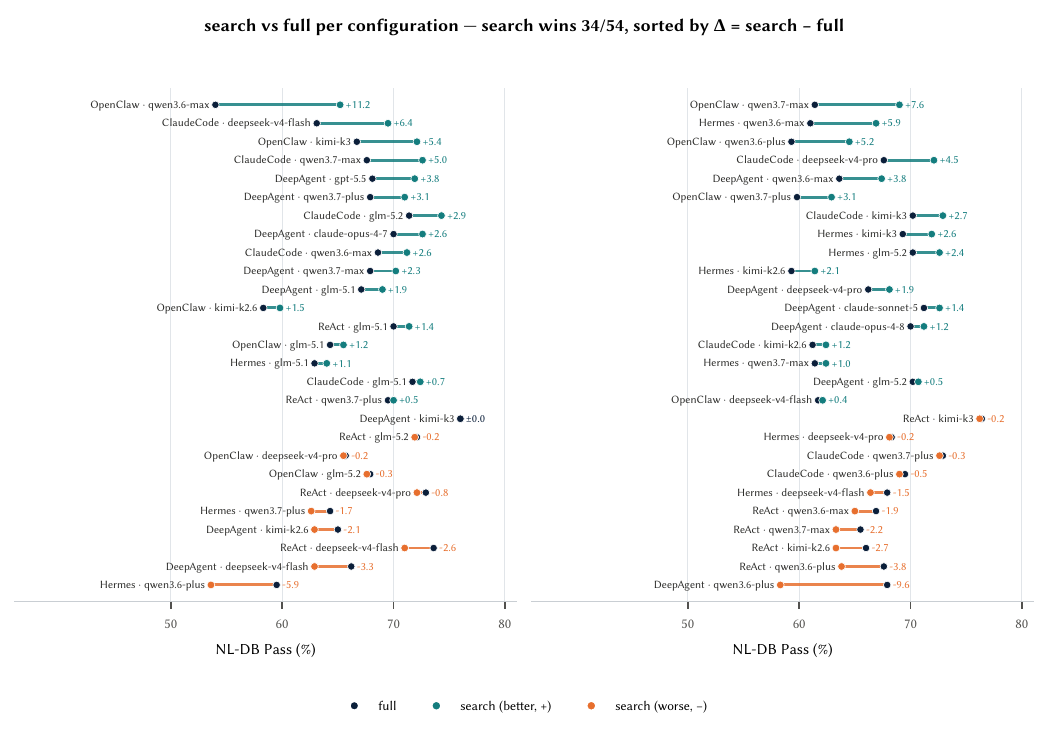}
  \caption{Full and search NL-DB values for the 54 combinations evaluated under both packages, sorted by the search-minus-full difference. Dark dots show full, teal dots show search where it is higher, orange dots where it is lower, and each row is annotated with its signed difference.}
  \Description{A two-column dumbbell chart. Each row prints a harness and model label beside a pair of dots connected by a line, a dark full-package dot and a colored search-package dot, teal when search is higher and orange when lower, with the signed difference printed at the right end of the pair.}
  \label{fig:e6b-interface-dumbbell}
\end{figure*}

OfficePrism presents every task through two complete interface packages, and Figure~\ref{fig:e6b-interface-dumbbell} compares them for each of the 54 harness--model combinations evaluated under both. Rows are sorted by the search-minus-full difference and dealt alternately into two columns so that both columns remain sorted; the dark dot is the full value, the colored dot the search value (teal when search is higher, orange when lower), and the signed difference is annotated at each pair. Hiding the native schemas behind discovery-and-dispatch is not the handicap one might expect: search is ahead for 34 of the 54 combinations with a median difference of $+1.2$ points. The gains concentrate where the full schema context is most burdensome---smaller or older models such as Qwen~3.6-max improve by up to $+11.2$ points---whereas the largest losses (down to $-9.6$ points) also occur for weaker models that fail to operate the discovery tools reliably. Overall, the search package performs significantly better than the full package.

\subsection{Effect of Tool-Pool Size}
\label{app:ext-tool-scaling}

\begin{figure*}[!t]
  \centering
  \includegraphics[width=0.8\textwidth]{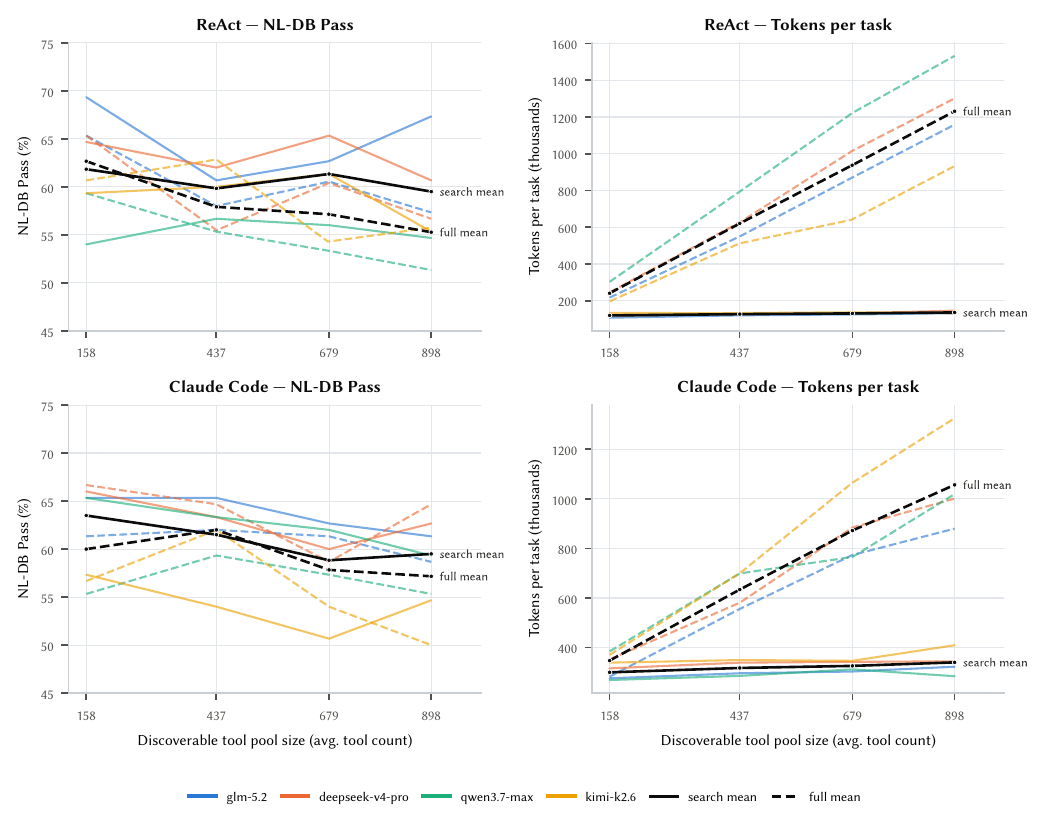}
  \caption{Tool-pool scaling on the auxiliary 150-task subset. Rows show the ReAct and Claude Code harnesses; columns show NL-DB Pass and tokens per task against the average discoverable tool count. Solid lines are the search package, dashed lines the full package, and bold black lines the package means.}
  \Description{Four line panels arranged two by two. The top row shows the ReAct harness and the bottom row the Claude Code harness. Left panels show NL-DB Pass and right panels show tokens per task in thousands, each with four model lines in solid search and dashed full styles plus bold black package means labeled search mean and full mean.}
  \label{fig:e6b-tool-scaling}
\end{figure*}

Real enterprise deployments expose far more tools than any single task needs, and Figure~\ref{fig:e6b-tool-scaling} measures what that costs. On an auxiliary 150-task subset, the discoverable pool is widened from one to all seven business domains (158 to 898 tools) while each task itself remains single-domain, for four models under the ReAct and Claude Code harnesses and both packages; cells backed by fewer than one hundred cases are excluded. Rows correspond to harnesses and columns to NL-DB Pass and tokens per task; solid lines are the search package, dashed lines the full package, and bold black lines the package means. The two packages diverge sharply. Under the full package, injecting every schema into context lets the mean token cost grow from 294k to 1{,}144k per task---nearly a factor of four---while accuracy falls by 5.1 points. Under the search package the same growth costs almost nothing: tokens stay nearly flat (206k to 237k) and accuracy declines by only 3.1 points. Discovery-and-dispatch therefore absorbs tool-pool growth that schema injection must pay for, which is the central practical argument for the search interface as catalogs scale.

\subsection{Failure Root-Cause Analysis}
\label{app:ext-error-causes}

\begin{figure*}[!t]
  \centering
  \includegraphics[width=0.73\textwidth]{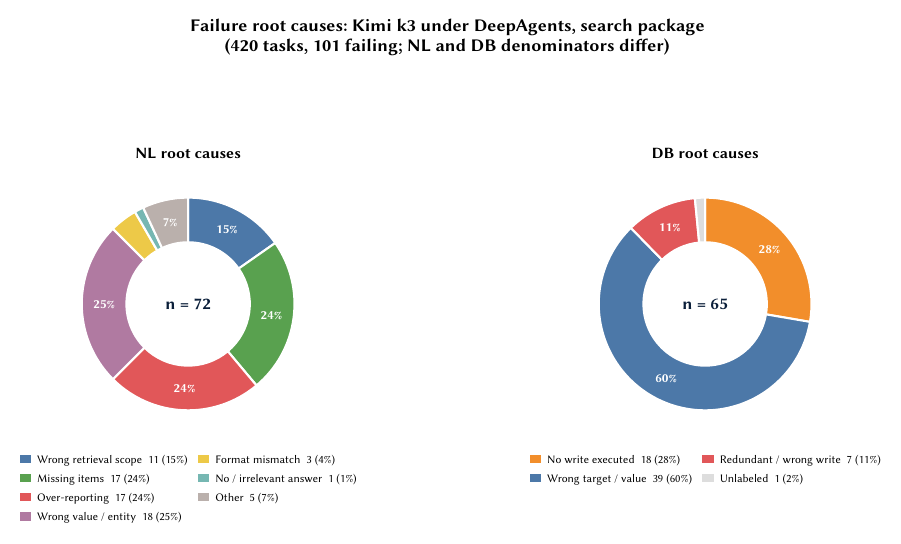}
  \caption{Failure root-cause distributions for the strongest leaderboard combination (Kimi~k3 under DeepAgents, search package) on the 420-task display set. The left donut shows the natural-language causes and the right donut the database causes, each labeled with per-cause counts and shares. A failing task can carry both an NL and a DB label, so the two denominators differ.}
  \Description{Two donut charts. The left donut shows natural-language root causes and the right donut database root causes for the search package. Each donut shows percentage labels on its wedges, the number of labeled failures in its center, and a legend with per-cause counts below.}
  \label{fig:e6b-error-causes}
\end{figure*}

Pass rates tell us how often agents fail, but not why. Figure~\ref{fig:e6b-error-causes} therefore looks inside the failures of the best combination on the leaderboard, Kimi~k3 under DeepAgents with the search package, which fails 101 of the 420 tasks. Every failed task is labeled with a cause on two sides: on the answer side we compare the agent's answer with the expected answer, and on the database side we compare the tool calls it made with the reference chain. A task can fail on both sides at once, so the two donuts have different totals. Two findings stand out. First, even the best system fails mostly on \emph{finding the right data}, not on wording: wrong values or entities (25\%), extra items (24\%), and missing items (24\%) make up nearly three quarters of the answer-side causes, while formatting problems are rare (4\%). The typical mistake is picking the wrong one of two similarly named records, or searching with a slightly wrong filter and returning too many or too few entries. Second, the database-side failures concentrate in one pattern: 60\% of them write to the wrong record or fill in the wrong value---the agent finds the object but writes the wrong content---and another 28\% never execute the required write at all; redundant or duplicate writes are rare (11\%). In short, the joint criterion does more than score tasks: it separates retrieval errors, value errors, and write-execution errors that a single pass rate would lump together, and it shows that the remaining headroom lies in precise record selection and careful write arguments rather than in language ability.

\subsection{Typical Failure Cases}
\label{app:ext-case-studies}

Beyond the aggregate root-cause distributions, we examine two representative failures in depth. Both come from evaluation trajectories of Claude Code with Kimi~k3 on this benchmark, and both were selected under three criteria: the failure was manually re-verified and is unambiguously attributable to the evaluated system (the reference answer and seed data are free of ambiguity and scoring disputes); each case corresponds to one distinct design pillar of the benchmark, showing that the failure mode is detectable only through that design; and each case was re-validated on the latest task set. Success is judged by the joint natural-language-and-database criterion, and every task runs on a fresh, isolated database. Figure~\ref{fig:e6b-failure-cases} summarizes the two cases; the paragraphs below analyze them.

The first case is a reasoning error that leaves no linguistic trace. The task is a conditional write: the agent must check the contract expiry dates of three employees involved in a fraud investigation and file a high-priority alert ticket listing exactly those whose contracts expire before 2025-12-31. The agent executes the workflow almost perfectly---it locates the case and the three employees, retrieves all three expiry dates, and even recovers from a rejected enum value to create the ticket---and its answer is fluent and confident. The single error is one date comparison: it treats 2025-10-29 as \emph{later} than 2025-12-31, so the condition branches the wrong way and the alert list omits an employee whose contract genuinely expires in time (while including, under its own improvised rule, an employee with no active contract). No fluency- or format-based evaluation would notice anything wrong; the conditional-write design turns this invisible reasoning slip into a machine-checkable difference in the database, and the NL and DB criteria both catch the wrong list. As an attribution check, another harness with the same model passes this task under the same conditions, confirming that the failure belongs to the evaluated system rather than to the task.

The second case fails while appearing to succeed at every visible step. The task is a two-step write: turn off the user's email auto-reply. The agent's object selection is impeccable---it lists the user's two mail accounts and correctly picks the active primary one over an archived alias---and it calls the right update tool. The failure happens in the last millimeter: it passes the parameter as the \emph{string} \texttt{"False"} instead of the boolean \texttt{False}. The call returns without error, and the agent confidently reports success, even quoting the returned field value and update timestamp as evidence; but a non-empty string is truthy, so the auto-reply effectively remains on---the opposite of what the user asked. Every part of the interaction that a user (or a call-success- or answer-based evaluation) can see looks correct; the error exists only in the type of a stored field. Only a type-sensitive check of the final database state separates ``said it was done'' from ``actually done.'' For deployment this is the most dangerous defect class: the user receives a clear success report and has no reason to double-check, while the system state has silently diverged from their belief.

Together, the two cases anchor two design pillars of the benchmark---conditional write tasks with joint NL/DB checking, and type-sensitive verification of the final database state---and illustrate that consequential agent failures concentrate precisely in the blind spots of conventional evaluation.

\begin{figure*}[!t]
  \centering
  \footnotesize
  \sbox0{\begin{minipage}[t]{0.465\textwidth}
    \textbf{Case 1: date-comparison reasoning error}\\[2pt]
    \textbf{Task}: I'm following up on that fraud report case (the one Zhang Qin submitted, severity high). Three people are involved---please check each one's current contract, mainly the expiry date. If anyone's contract expires before 2025-12-31, file an HR service ticket titled `Contract-expiry alert for employees involved in a compliance investigation', note in the description who is expiring and on which date, and set the priority to high. If nobody is at risk, don't file anything---just tell me the result.\\[3pt]
    \cmark~Locates the report case and the three involved employees\\
    \cmark~Retrieves the expiry dates: 2025-08-14 / no active contract / 2025-10-29\\
    \cmark~First ticket call hits an enum error; retries and creates the alert ticket\\
    \xmark~\textbf{Treats 2025-10-29 as ``later than'' 2025-12-31, so the alert list omits an employee}\\[3pt]
    \textbf{Verdict:} NL \xmark~~DB \xmark
  \end{minipage}}%
  \fbox{\usebox0}\hfill
  \fbox{\begin{minipage}[t][\dimexpr\ht0+\dp0\relax][t]{0.465\textwidth}
    \textbf{Case 2: boolean parameter written as a string}\\[2pt]
    \textbf{Task}: Please turn off my email auto-reply---I set it before my leave and I'm back now.\\[3pt]
    \cmark~Lists the user's two mail accounts and picks the active primary one (skips the archived alias)\\
    \cmark~Calls the settings-update tool; the call returns without error and the agent reports success, quoting the returned record\\
    \xmark~\textbf{Passes \texttt{auto\_reply\_enabled} as the string \texttt{"False"} instead of boolean \texttt{False}; a non-empty string is truthy, so the auto-reply stays on}\\[3pt]
    \textbf{Verdict:} NL \cmark~~DB \xmark
  \end{minipage}}
  \caption{Two representative failure cases of Claude Code with Kimi~k3. Left: a single wrong date comparison inside an otherwise flawless conditional-write workflow. Right: a write that succeeds at every visible step but stores a string where a boolean is required, leaving the final state wrong while the agent reports success.}
  \Description{Two framed text panels. The left panel summarizes a contract-expiry alert task in which the agent performs every step correctly except one date comparison, which causes the alert list to omit an employee; both the natural-language and database checks fail. The right panel summarizes an email auto-reply task in which the agent selects the correct account and calls the correct tool but passes the value False as a string, so the auto-reply remains enabled and only the database check fails.}
  \label{fig:e6b-failure-cases}
\end{figure*}

\input{sections/rebuttal_add}
\section{Implementation and Result Details}
\label{sec:artifact-details}

All reported evaluations use the same checksummed 420-task JSON. The artifact
released upon acceptance will include the task file, runner, evaluator,
interface documentation, experiment configurations, licenses, and checksums.

\subsection{Construction Configuration}
\label{app:construction-config}

Tables~\ref{tab:construction-config} and~\ref{tab:construction-details}
summarize the implemented construction path.

The survey, generator, and quality-filter prompts are written in Chinese. The
construction inputs include the business-rule files and 21 few-shot records.
English translations are provided for documentation; model calls use the
Chinese prompts.

\begin{table}[htb]
\caption{Grounding and synthesis configuration.}
\label{tab:construction-config}
\centering
\footnotesize
\setlength{\tabcolsep}{2pt}
\begin{tabular}{@{}>{\raggedright\arraybackslash}p{1.55cm}>{\raggedright\arraybackslash}p{5.25cm}@{}}
\toprule
Component & Implementation \\
\midrule
Scenario inventory & A cached survey is used when available. Otherwise an LLM groups tools and tables, with a full-domain parse fallback. L1--L2 use one group and L3 merges up to three same-domain groups. \\
Grounding packet & The packet contains an active caller, tool names and parameters, at most 300 documentation characters per tool, six shared tables, scenario tables, and at most ten prefix records per included table. \\
Prompt and example & Generation uses at most one level-matched example, preferring same-domain full coverage, then same-domain positive overlap, then cross-function full coverage. \\
\bottomrule
\end{tabular}
\end{table}

\begin{table}[htb]
\caption{Model calls, automated checks, and task records.}
\label{tab:construction-details}
\centering
\footnotesize
\setlength{\tabcolsep}{2pt}
\begin{tabular}{@{}>{\raggedright\arraybackslash}p{1.55cm}>{\raggedright\arraybackslash}p{5.25cm}@{}}
\toprule
Component & Implementation \\
\midrule
Model calls & Generation and filtering are separate calls that share one model-configuration object. \\
Randomness and attempts & Caller, example, and L3 group choices use a process-wide RNG. UUID4 supplies task IDs. Quota filling has at most three rounds and three attempts per scheduled item. \\
Replay and filter & Replay executes the stored chain sequentially on a copied toolkit. Logic, naturalness, and level-alignment scores must each reach four. \\
Redundancy & Structural collapse uses reference-tool set, caller role, and level. Text similarity uses caller-role--level buckets at 0.85; character trigrams are the default encoder. \\
Task record & OfficePrism-420 stores caller, domain, level, group, tool and table subsets, automated-check fields, a reference tool chain, and an expected answer. \\
\bottomrule
\end{tabular}
\end{table}

\subsection{Selection and Environment Statistics}
\label{app:selection-provenance}

Table~\ref{tab:artifact-lifecycle} separates construction artifacts from the
benchmark set. The 574-task count comes from the supplied curation note.
The supplied JSON defines the 420-task benchmark and contains 420 unique
identifiers across all seven business functions, with L1/L2/L3 counts of
105/140/175.
Figure~\ref{fig:task-profile} reports the function-by-level composition.

\begin{table}[htb]
\caption{Construction lifecycle and benchmark identity.}
\label{tab:artifact-lifecycle}
\centering
\footnotesize
\setlength{\tabcolsep}{3pt}
\begin{tabular}{@{}>{\raggedright\arraybackslash}p{2.2cm}r>{\raggedright\arraybackslash}p{4.15cm}@{}}
\toprule
Artifact stage & $N$ & Evidence source \\
\midrule
Machine-check survivors & 1,135 & Saved single-domain generation records. \\
After redundancy control & 852 & Cleaned single-domain construction files. \\
Human-curated pool & 574 & Aggregate reported by the supplied curation note. \\
OfficePrism-420 & 420 & Checksummed task JSON used by the evaluator and every reported configuration; L1/L2/L3 = 105/140/175. JSON SHA-256 (abbrev.) \texttt{7d041782c1ae...7b0096d8}. \\
\bottomrule
\end{tabular}
\end{table}

\textbf{Task-set selection.}
 The supplied metadata describes OfficePrism-420 as a fixed 300-task base plus
120 additions: 86 from one candidate source and 34 from another. Candidate
selection uses minimax pilot performance and, on ties, prefers
search-friendly cases. The resulting set contains 105 L1, 140 L2, and 175 L3
tasks across seven functions.

\textbf{Task-record profile.}
Table~\ref{tab:task-record-profile} reports descriptors recomputed directly
from the checksummed JSON. Ticket and response lengths count Unicode code
points in the question and expected-response fields; witness length counts
parsed calls in \texttt{tool\_chain}.

\input{generated/task_record_profile.tex}

The task set contains 264 distinct construction scenario-group labels and
1,965 reference-witness calls over 408 distinct function names. Mean
construction-side subset sizes are 54.13 tool names and 17.17 table names per
task. Scenario groups and construction subsets organize the inputs supplied to
task synthesis.

\textbf{Task-local environment scale.}
Each active runtime merges the common organizational base with exactly one
function overlay. Table~\ref{tab:environment-scale} separates overlay state
from the resulting task-local state. The shared infrastructure contains 16
collections, 782 records, 27 READ functions, 4 WRITE functions, and 2 generic
functions. Across task-local environments, the active state ranges from 42 to
61 collections and from 4,701 to 17,126 records, with 135--198 registered
functions.

\begin{table}[htb]
\caption{Task-local environment inventory.}
\label{tab:environment-scale}
\centering
\footnotesize
\setlength{\tabcolsep}{2.5pt}
\begin{tabular}{@{}>{\raggedright\arraybackslash}p{1.85cm}ccc@{}}
\toprule
Layer or function & \shortstack{Overlay state\\Coll. / records} & \shortstack{Active state\\Coll. / records} & \shortstack{Registered functions\\READ / WRITE / generic} \\
\midrule
Shared infrastructure & --- & 16 / 782 & 27 / 4 / 2 \\
Administration & 42 / 16,344 & 58 / 17,126 & 105 / 47 / 2 \\
Finance & 26 / 4,334 & 42 / 5,116 & 82 / 56 / 2 \\
Human resources & 44 / 11,381 & 60 / 12,163 & 115 / 36 / 2 \\
Information technology & 45 / 6,357 & 61 / 7,139 & 118 / 78 / 2 \\
Legal affairs & 37 / 4,105 & 53 / 4,887 & 110 / 66 / 2 \\
Procurement & 29 / 3,919 & 45 / 4,701 & 86 / 53 / 2 \\
Project work & 27 / 9,310 & 43 / 10,092 & 84 / 49 / 2 \\
\bottomrule
\end{tabular}
\end{table}

\textbf{Synthetic organization.}
Table~\ref{tab:synthetic-population} reports selected attributes of the shared
organization. Employee and profile rows use all 75 employee records; recorded
nationality uses the 20 visa records in the human-resources overlay.

\begin{table}[htb]
\caption{Synthetic-organization profile.}
\label{tab:synthetic-population}
\centering
\footnotesize
\setlength{\tabcolsep}{3pt}
\begin{tabular}{@{}>{\raggedright\arraybackslash}p{2.65cm}>{\centering\arraybackslash}p{0.8cm}>{\raggedright\arraybackslash}p{3.9cm}@{}}
\toprule
Source and attribute & $N$ & Value \\
\midrule
\texttt{base.employee}\\Worker type & 75 & 60 full-time, 5 contractors, 5 interns, 3 campus recruits, and 2 campus interns \\
\texttt{base.employee}\\Employment status & 75 & 63 active, 7 probation, 3 terminated, and 2 on leave \\
\texttt{base.employee}\\Direct manager & 75 & Reference present for 74 employees \\
Shared organization\\Entity counts & --- & 12 departments, 22 teams, 28 positions, and 8 roles \\
\texttt{base.user\_profile}\\Preferred locale & 75 & 43 \texttt{zh-CN}, 19 \texttt{ja-JP}, and 13 \texttt{en-US} \\
\texttt{base.user\_profile}\\Timezone & 75 & 51 Shanghai, 13 Chongqing, and 11 Hong Kong \\
\texttt{hr.visa\_record}\\Recorded nationality & 20 & 5 US, 4 CA, 4 JP, 3 KR, 3 UK, and 1 SG \\
\bottomrule
\end{tabular}
\end{table}

\subsection{Runner Outcome Fields}
\label{app:legacy-metrics}

The runner stores separate per-case component values and two convenience
fields inherited from the implementation. \texttt{success} thresholds the
natural-language component for tasks with an expected answer, whereas
\texttt{strict\_success} thresholds the runner's multiplicative reward. They
are included in the full result ledger alongside NL--DB and recorded steps.

Task-local reset permits the state evaluator to reconstruct the agent outcome
and the reference-replay outcome from the same initial environment. Witness
replay provides the comparison state, while NL--DB combines the
expected-outcome judgment with the reconstructed final-state check.

\subsection{Displayed OfficePrism Result Ledger}
\label{app:displayed-results}

Tables~\ref{tab:e6-full-react}--\ref{tab:e6-full-hermes} disclose the 124 displayed rows underlying
Figure~\ref{fig:e6-main-results}, grouped by harness. Every row aggregates
case-level outcomes over the same 420-task artifact, whose fixed level
denominators are 105 L1, 140 L2, and 175 L3 tasks.

NL--DB is the headline field. L1--L3 are its level-specific
rates; Success, Strict, and Steps preserve the remaining displayed
diagnostics. At the supplied precision, Success equals 100 times average NL
and Strict equals 100 times average reward in all rows, so the duplicated
decimal columns are omitted. The manuscript source includes the exact task and
leaderboard hashes and the generated CSV.

\input{generated/e6_full_results_appendix}

\textbf{Case accounting.}
Each overall rate has denominator 420, so one task changes an unrounded rate by
\(100/420=0.238\) percentage points. The corresponding L1, L2, and L3
resolutions are \(100/105=0.952\), \(100/140=0.714\), and
\(100/175=0.571\) points. For every displayed row, weighting its three
one-decimal level rates by 105/140/175 reproduces the reported overall NL--DB
rate within 0.075 points. This check is consistent with per-case aggregation
over the advertised fixed subsets.

\textbf{Free-text DB evaluation.}
For the 104 cases whose reference calls contain Chinese free text, DB status
uses the structural and text comparison described in
Section~\ref{sec:evaluation}. The result ledger below contains all 124
configurations.

\FloatBarrier

%% file: sections/rebuttal_add.tex

\subsection{Base-set Comparison}
\label{subsec:base-set}

The benchmark cases were synthesized in two batches. To ensure that the
tool-binding mode remains consistent across the two syntheses, we verified
the results of both sets under the two tool-binding modes, as reported in
Table~\ref{tab:base-set}.

\begin{table}[H]
  \caption{Search-minus-full differences (percentage points) and search
  wins/ties/losses on the 300-task base and the 420-task set.}
  \label{tab:base-set}
  \centering
  \footnotesize
  \setlength{\tabcolsep}{3pt}
  \begin{tabular}{@{}lccc@{}}
    \toprule
    Task set & Median & Mean & Wins/ties/losses \\
    \midrule
    300-task base & $+0.67$ pp & $+0.71$ pp & 30/3/17 \\
    420-task set & $+1.07$ pp & $+1.05$ pp & 30/1/19 \\
    \bottomrule
  \end{tabular}
\end{table}

The two sets show no material difference, and the conclusion that
\texttt{search} outperforms \texttt{full} remains unchanged.

\subsection{Judge Validation}
\label{subsec:judge-validation}

We rescored the same saved outputs across all 124 configurations $\times$ 420
tasks using three judges, with three passes each
(Table~\ref{tab:judge-passes}).

\begin{table}[H]
  \caption{Mean joint NL--DB scores from three judges, three passes each.}
  \label{tab:judge-passes}
  \centering
  \footnotesize
  \begin{tabular}{@{}lccc@{}}
    \toprule
    Judge & Pass 1 & Pass 2 & Pass 3 \\
    \midrule
    \texttt{qwen3-max} & 66.97\% & 67.00\% & 67.00\% \\
    \texttt{qwen3.8-max} & 67.24\% & 67.26\% & 67.25\% \\
    \texttt{kimi-k3} & 67.24\% & 67.24\% & 67.24\% \\
    \bottomrule
  \end{tabular}
\end{table}

In addition, the NL judge achieves \textbf{98.7\% agreement with human
assessments}, and \textbf{the core models' aggregate ordering remains
unchanged across the three judge models}. In a 150-case audit of
DeepAgents-kimi-k3-search, \texttt{qwen3-max} agreed with human
assessments in 148 cases.

\subsection{Stability Analysis}
\label{subsec:stability}

We added two execution rounds covering 24 configurations across three
harnesses (ReAct, DeepAgents, and Hermes), four models, and both interfaces,
each on all 420 tasks. \textbf{The four models retain the same aggregate
ordering in all three runs.} Together with the rescored original run, this
yields 30,240 task evaluations, including 20,160 newly executed evaluations
(Table~\ref{tab:stability-runs}).

\begin{table}[H]
  \caption{NL--DB scores of the four models across the three runs.}
  \label{tab:stability-runs}
  \centering
  \footnotesize
  \begin{tabular}{@{}lccc@{}}
    \toprule
    Model & Original run & Added run 1 & Added run 2 \\
    \midrule
    \texttt{kimi-k3} & 74.4\% & 73.1\% & 72.0\% \\
    \texttt{glm-5.2} & 71.3\% & 70.2\% & 69.1\% \\
    \texttt{qwen3.7-max} & 65.3\% & 65.0\% & 65.3\% \\
    \texttt{qwen3.6-plus} & 61.8\% & 61.3\% & 62.0\% \\
    \bottomrule
  \end{tabular}
\end{table}

%% file: generated/task_record_profile.tex
\begin{table}[htb]
\caption{Task-record profile (mean / median).}
\label{tab:task-record-profile}
\centering
\footnotesize
\setlength{\tabcolsep}{3.2pt}
\begin{tabular*}{\columnwidth}{@{\extracolsep{\fill}}lrrrr@{}}
\toprule
Level & $N$ & Ticket & Witness & Response \\
\midrule
L1 & 105 & 38.7/35 & 1.50/1 & 129.8/69 \\
L2 & 140 & 91.8/87.5 & 3.34/3 & 146.5/122 \\
L3 & 175 & 209.0/200 & 7.66/7 & 390.1/364 \\
All & 420 & 127.4/103.5 & 4.68/4 & 243.8/163 \\
\bottomrule
\end{tabular*}
\end{table}

%% file: generated/e6_full_results_appendix.tex
\begin{table*}[!tb]
\caption{OfficePrism-420 aggregates for ReAct display labels.}
\label{tab:e6-full-react}
\centering
\footnotesize
\setlength{\tabcolsep}{3.1pt}
\renewcommand{\arraystretch}{1.08}
\begin{tabular*}{\textwidth}{@{\extracolsep{\fill}}>{\raggedright\arraybackslash}p{3.0cm}lrrrrrrr@{}}
\toprule
Model & Pkg. & NL--DB & L1 & L2 & L3 & Success & Strict & Steps \\
\midrule
kimi-k3 & \texttt{full} & 76.4 & 88.6 & 82.1 & 64.6 & 85.5 & 61.4 & 13.1 \\
kimi-k3 & \texttt{search} & 76.2 & 90.5 & 78.6 & 65.7 & 83.3 & 58.3 & 19.0 \\
claude-opus-4-7 & \texttt{search} & 74.0 & 89.5 & 74.3 & 64.6 & 80.5 & 57.9 & 17.3 \\
claude-opus-4-8 & \texttt{search} & 74.0 & 92.4 & 72.9 & 64.0 & 81.0 & 57.9 & 17.2 \\
deepseek-v4-flash & \texttt{full} & 73.6 & 86.7 & 77.1 & 62.9 & 81.4 & 59.3 & 14.8 \\
claude-sonnet-5 & \texttt{search} & 73.1 & 88.6 & 79.3 & 58.9 & 80.2 & 58.3 & 20.5 \\
deepseek-v4-pro & \texttt{full} & 72.9 & 86.7 & 78.6 & 60.0 & 81.0 & 59.5 & 13.3 \\
deepseek-v4-pro & \texttt{search} & 72.1 & 85.7 & 80.7 & 57.1 & 80.2 & 58.1 & 19.4 \\
glm-5.2 & \texttt{full} & 72.1 & 86.7 & 75.0 & 61.1 & 80.7 & 57.6 & 11.8 \\
glm-5.2 & \texttt{search} & 71.9 & 86.7 & 75.0 & 60.6 & 79.0 & 58.1 & 19.8 \\
glm-5.1 & \texttt{search} & 71.4 & 85.7 & 75.7 & 59.4 & 78.3 & 57.4 & 19.1 \\
deepseek-v4-flash & \texttt{search} & 71.0 & 84.8 & 78.6 & 56.6 & 77.9 & 56.2 & 23.4 \\
gpt-5.5 & \texttt{search} & 71.0 & 86.7 & 77.1 & 56.6 & 77.4 & 58.1 & 22.4 \\
glm-5.1 & \texttt{full} & 70.0 & 86.7 & 75.7 & 55.4 & 78.8 & 56.9 & 12.1 \\
qwen3.7-plus & \texttt{search} & 70.0 & 85.7 & 73.6 & 57.7 & 77.4 & 52.9 & 20.6 \\
qwen3.7-plus & \texttt{full} & 69.5 & 82.9 & 69.3 & 61.7 & 78.8 & 53.8 & 12.7 \\
qwen3.6-plus & \texttt{full} & 67.6 & 81.9 & 67.9 & 58.9 & 75.5 & 51.9 & 12.5 \\
qwen3.6-max & \texttt{full} & 66.9 & 81.9 & 65.7 & 58.9 & 75.0 & 52.4 & 11.7 \\
kimi-k2.6 & \texttt{full} & 66.0 & 84.8 & 77.9 & 45.1 & 65.5 & 51.4 & 12.0 \\
qwen3.7-max & \texttt{full} & 65.5 & 81.9 & 62.1 & 58.3 & 72.4 & 50.2 & 11.5 \\
qwen3.6-max & \texttt{search} & 65.0 & 78.1 & 70.7 & 52.6 & 71.4 & 48.8 & 19.7 \\
qwen3.6-plus & \texttt{search} & 63.8 & 76.2 & 66.4 & 54.3 & 69.8 & 48.6 & 20.6 \\
kimi-k2.6 & \texttt{search} & 63.3 & 81.9 & 67.9 & 48.6 & 64.3 & 47.9 & 18.8 \\
qwen3.7-max & \texttt{search} & 63.3 & 80.0 & 65.0 & 52.0 & 70.7 & 47.6 & 18.3 \\
\bottomrule
\end{tabular*}
\end{table*}

\begin{table*}[!tb]
\caption{OfficePrism-420 aggregates for DeepAgents display labels.}
\label{tab:e6-full-deepagents}
\centering
\footnotesize
\setlength{\tabcolsep}{3.1pt}
\renewcommand{\arraystretch}{1.08}
\begin{tabular*}{\textwidth}{@{\extracolsep{\fill}}>{\raggedright\arraybackslash}p{3.0cm}lrrrrrrr@{}}
\toprule
Model & Pkg. & NL--DB & L1 & L2 & L3 & Success & Strict & Steps \\
\midrule
kimi-k3 & \texttt{full} & 76.0 & 91.4 & 82.9 & 61.1 & 82.1 & 58.8 & 7.5 \\
kimi-k3 & \texttt{search} & 76.0 & 91.4 & 82.1 & 61.7 & 82.4 & 57.4 & 11.6 \\
claude-opus-4-7 & \texttt{search} & 72.6 & 92.4 & 71.4 & 61.7 & 79.0 & 55.5 & 10.3 \\
claude-sonnet-5 & \texttt{search} & 72.6 & 93.3 & 76.4 & 57.1 & 79.8 & 58.6 & 13.0 \\
gpt-5.5 & \texttt{search} & 71.9 & 85.7 & 78.6 & 58.3 & 79.0 & 57.6 & 12.3 \\
claude-opus-4-8 & \texttt{search} & 71.2 & 86.7 & 74.3 & 59.4 & 78.6 & 55.5 & 10.5 \\
claude-sonnet-5 & \texttt{full} & 71.2 & 91.4 & 72.9 & 57.7 & 78.3 & 58.1 & 7.5 \\
qwen3.7-plus & \texttt{search} & 71.0 & 85.7 & 74.3 & 59.4 & 76.9 & 53.8 & 11.4 \\
glm-5.2 & \texttt{search} & 70.7 & 88.6 & 73.6 & 57.7 & 78.6 & 55.7 & 12.2 \\
glm-5.2 & \texttt{full} & 70.2 & 83.8 & 75.7 & 57.7 & 79.0 & 56.0 & 7.8 \\
qwen3.7-max & \texttt{search} & 70.2 & 82.9 & 75.7 & 58.3 & 76.4 & 54.0 & 10.7 \\
claude-opus-4-7 & \texttt{full} & 70.0 & 89.5 & 71.4 & 57.1 & 76.7 & 56.4 & 6.7 \\
claude-opus-4-8 & \texttt{full} & 70.0 & 89.5 & 72.1 & 56.6 & 74.0 & 58.1 & 7.0 \\
glm-5.1 & \texttt{search} & 69.0 & 88.6 & 72.1 & 54.9 & 77.1 & 54.8 & 12.0 \\
deepseek-v4-pro & \texttt{search} & 68.1 & 81.9 & 76.4 & 53.1 & 74.3 & 53.1 & 13.1 \\
gpt-5.5 & \texttt{full} & 68.1 & 84.8 & 72.1 & 54.9 & 75.7 & 56.4 & 7.8 \\
qwen3.6-plus & \texttt{full} & 67.9 & 78.1 & 68.6 & 61.1 & 76.0 & 51.2 & 7.2 \\
qwen3.7-max & \texttt{full} & 67.9 & 83.8 & 68.6 & 57.7 & 74.5 & 51.7 & 6.8 \\
qwen3.7-plus & \texttt{full} & 67.9 & 79.0 & 73.6 & 56.6 & 75.7 & 51.4 & 7.2 \\
qwen3.6-max & \texttt{search} & 67.4 & 84.8 & 67.1 & 57.1 & 72.6 & 50.5 & 11.1 \\
glm-5.1 & \texttt{full} & 67.1 & 81.9 & 71.4 & 54.9 & 75.0 & 53.6 & 7.3 \\
deepseek-v4-flash & \texttt{full} & 66.2 & 80.0 & 70.7 & 54.3 & 74.8 & 53.6 & 8.4 \\
deepseek-v4-pro & \texttt{full} & 66.2 & 81.9 & 72.9 & 51.4 & 74.8 & 53.6 & 8.5 \\
kimi-k2.6 & \texttt{full} & 65.0 & 80.0 & 71.4 & 50.9 & 71.9 & 51.4 & 7.4 \\
qwen3.6-max & \texttt{full} & 63.6 & 81.0 & 62.9 & 53.7 & 70.7 & 49.8 & 6.8 \\
deepseek-v4-flash & \texttt{search} & 62.9 & 84.8 & 75.0 & 40.0 & 69.3 & 50.7 & 12.2 \\
kimi-k2.6 & \texttt{search} & 62.9 & 83.8 & 65.7 & 48.0 & 70.5 & 51.2 & 11.7 \\
qwen3.6-plus & \texttt{search} & 58.3 & 69.5 & 61.4 & 49.1 & 65.0 & 45.5 & 12.6 \\
\bottomrule
\end{tabular*}
\end{table*}

\begin{table*}[!tb]
\caption{OfficePrism-420 aggregates for Claude Code display labels.}
\label{tab:e6-full-claude-code}
\centering
\footnotesize
\setlength{\tabcolsep}{3.1pt}
\renewcommand{\arraystretch}{1.08}
\begin{tabular*}{\textwidth}{@{\extracolsep{\fill}}>{\raggedright\arraybackslash}p{3.0cm}lrrrrrrr@{}}
\toprule
Model & Pkg. & NL--DB & L1 & L2 & L3 & Success & Strict & Steps \\
\midrule
glm-5.2 & \texttt{search} & 74.3 & 87.6 & 82.1 & 60.0 & 82.4 & 61.2 & 11.3 \\
kimi-k3 & \texttt{search} & 72.9 & 84.8 & 80.7 & 59.4 & 80.0 & 58.8 & 12.1 \\
qwen3.7-plus & \texttt{full} & 72.9 & 90.5 & 75.0 & 60.6 & 80.2 & 59.5 & 7.6 \\
qwen3.7-max & \texttt{search} & 72.6 & 86.7 & 80.0 & 58.3 & 80.0 & 59.8 & 11.7 \\
qwen3.7-plus & \texttt{search} & 72.6 & 90.5 & 77.1 & 58.3 & 79.3 & 59.0 & 12.2 \\
claude-opus-4-8 & \texttt{search} & 72.4 & 87.6 & 76.4 & 60.0 & 78.3 & 55.7 & 10.2 \\
glm-5.1 & \texttt{search} & 72.4 & 88.6 & 81.4 & 55.4 & 80.2 & 60.0 & 12.5 \\
deepseek-v4-pro & \texttt{search} & 72.1 & 84.8 & 79.3 & 58.9 & 77.6 & 57.6 & 12.4 \\
claude-opus-4-7 & \texttt{search} & 71.9 & 87.6 & 75.7 & 59.4 & 77.6 & 56.9 & 10.0 \\
glm-5.1 & \texttt{full} & 71.7 & 88.6 & 76.4 & 57.7 & 80.2 & 57.9 & 8.1 \\
glm-5.2 & \texttt{full} & 71.4 & 83.8 & 75.7 & 60.6 & 79.8 & 57.6 & 7.6 \\
qwen3.6-max & \texttt{search} & 71.2 & 87.6 & 75.7 & 57.7 & 77.9 & 56.9 & 11.5 \\
kimi-k3 & \texttt{full} & 70.2 & 86.7 & 73.6 & 57.7 & 77.4 & 57.6 & 7.9 \\
deepseek-v4-flash & \texttt{search} & 69.5 & 83.8 & 75.0 & 56.6 & 76.0 & 56.0 & 12.7 \\
qwen3.6-plus & \texttt{full} & 69.5 & 87.6 & 76.4 & 53.1 & 76.4 & 56.2 & 7.7 \\
qwen3.6-plus & \texttt{search} & 69.0 & 82.9 & 75.7 & 55.4 & 75.0 & 55.5 & 12.2 \\
claude-sonnet-5 & \texttt{search} & 68.8 & 86.7 & 75.0 & 53.1 & 76.2 & 54.5 & 11.2 \\
qwen3.6-max & \texttt{full} & 68.6 & 81.9 & 69.3 & 60.0 & 75.2 & 56.2 & 7.7 \\
deepseek-v4-pro & \texttt{full} & 67.6 & 87.6 & 72.1 & 52.0 & 75.2 & 57.1 & 8.4 \\
qwen3.7-max & \texttt{full} & 67.6 & 84.8 & 70.0 & 55.4 & 74.8 & 55.5 & 8.4 \\
deepseek-v4-flash & \texttt{full} & 63.1 & 75.2 & 67.1 & 52.6 & 71.7 & 53.8 & 8.7 \\
gpt-5.5 & \texttt{search} & 62.4 & 77.1 & 69.3 & 48.0 & 67.1 & 51.9 & 12.6 \\
kimi-k2.6 & \texttt{search} & 62.4 & 88.6 & 68.6 & 41.7 & 62.6 & 50.0 & 11.3 \\
kimi-k2.6 & \texttt{full} & 61.2 & 81.9 & 67.9 & 43.4 & 60.5 & 48.3 & 8.0 \\
\bottomrule
\end{tabular*}
\end{table*}

\begin{table*}[!tb]
\caption{OfficePrism-420 aggregates for OpenClaw display labels.}
\label{tab:e6-full-openclaw}
\centering
\footnotesize
\setlength{\tabcolsep}{3.1pt}
\renewcommand{\arraystretch}{1.08}
\begin{tabular*}{\textwidth}{@{\extracolsep{\fill}}>{\raggedright\arraybackslash}p{3.0cm}lrrrrrrr@{}}
\toprule
Model & Pkg. & NL--DB & L1 & L2 & L3 & Success & Strict & Steps \\
\midrule
kimi-k3 & \texttt{search} & 72.1 & 90.5 & 77.1 & 57.1 & 76.9 & 57.1 & 13.2 \\
qwen3.7-max & \texttt{search} & 69.0 & 89.5 & 70.0 & 56.0 & 74.8 & 55.7 & 12.7 \\
glm-5.2 & \texttt{full} & 67.9 & 85.7 & 70.7 & 54.9 & 74.0 & 55.5 & 9.0 \\
glm-5.2 & \texttt{search} & 67.6 & 88.6 & 72.9 & 50.9 & 76.2 & 56.2 & 13.5 \\
kimi-k3 & \texttt{full} & 66.7 & 81.9 & 70.0 & 54.9 & 72.4 & 54.3 & 7.8 \\
deepseek-v4-pro & \texttt{full} & 65.7 & 83.8 & 64.3 & 56.0 & 71.7 & 52.4 & 8.8 \\
deepseek-v4-pro & \texttt{search} & 65.5 & 79.0 & 72.9 & 51.4 & 71.2 & 53.3 & 15.0 \\
glm-5.1 & \texttt{search} & 65.5 & 84.8 & 70.7 & 49.7 & 71.0 & 52.1 & 14.3 \\
qwen3.6-max & \texttt{search} & 65.2 & 81.9 & 68.6 & 52.6 & 71.2 & 51.9 & 13.3 \\
qwen3.6-plus & \texttt{search} & 64.5 & 83.8 & 68.6 & 49.7 & 71.2 & 50.7 & 12.8 \\
glm-5.1 & \texttt{full} & 64.3 & 75.2 & 70.7 & 52.6 & 70.7 & 50.0 & 8.5 \\
claude-opus-4-7 & \texttt{search} & 63.3 & 74.3 & 62.1 & 57.7 & 67.9 & 47.4 & 10.3 \\
qwen3.7-plus & \texttt{search} & 62.9 & 81.0 & 66.4 & 49.1 & 69.8 & 51.0 & 12.4 \\
deepseek-v4-flash & \texttt{search} & 62.1 & 76.2 & 69.3 & 48.0 & 67.9 & 51.4 & 14.1 \\
deepseek-v4-flash & \texttt{full} & 61.7 & 74.3 & 63.6 & 52.6 & 69.0 & 49.0 & 8.9 \\
claude-opus-4-8 & \texttt{search} & 61.4 & 72.4 & 62.9 & 53.7 & 66.2 & 47.6 & 10.7 \\
qwen3.7-max & \texttt{full} & 61.4 & 70.5 & 65.0 & 53.1 & 67.6 & 49.3 & 7.0 \\
kimi-k2.6 & \texttt{search} & 59.8 & 86.7 & 63.6 & 40.6 & 59.5 & 45.2 & 12.1 \\
qwen3.7-plus & \texttt{full} & 59.8 & 76.2 & 59.3 & 50.3 & 66.7 & 47.1 & 6.8 \\
qwen3.6-plus & \texttt{full} & 59.3 & 69.5 & 65.7 & 48.0 & 66.9 & 46.7 & 7.9 \\
kimi-k2.6 & \texttt{full} & 58.3 & 81.0 & 62.1 & 41.7 & 58.1 & 44.5 & 8.2 \\
qwen3.6-max & \texttt{full} & 54.0 & 71.4 & 55.7 & 42.3 & 62.1 & 46.0 & 8.0 \\
gpt-5.5 & \texttt{search} & 32.9 & 51.4 & 29.3 & 24.6 & 34.8 & 26.7 & 8.3 \\
claude-sonnet-5 & \texttt{search} & 22.6 & 31.4 & 17.9 & 21.1 & 24.8 & 18.3 & 4.8 \\
\bottomrule
\end{tabular*}
\end{table*}

\begin{table*}[!tb]
\caption{OfficePrism-420 aggregates for Hermes display labels.}
\label{tab:e6-full-hermes}
\centering
\footnotesize
\setlength{\tabcolsep}{3.1pt}
\renewcommand{\arraystretch}{1.08}
\begin{tabular*}{\textwidth}{@{\extracolsep{\fill}}>{\raggedright\arraybackslash}p{3.0cm}lrrrrrrr@{}}
\toprule
Model & Pkg. & NL--DB & L1 & L2 & L3 & Success & Strict & Steps \\
\midrule
glm-5.2 & \texttt{search} & 72.6 & 86.7 & 74.3 & 62.9 & 78.8 & 57.9 & 12.7 \\
kimi-k3 & \texttt{search} & 71.9 & 83.8 & 76.4 & 61.1 & 76.9 & 55.0 & 13.3 \\
glm-5.2 & \texttt{full} & 70.2 & 86.7 & 69.3 & 61.1 & 78.3 & 55.7 & 8.3 \\
claude-opus-4-7 & \texttt{search} & 70.0 & 83.8 & 72.9 & 59.4 & 74.8 & 54.0 & 10.3 \\
claude-opus-4-8 & \texttt{search} & 69.5 & 81.0 & 72.1 & 60.6 & 73.8 & 53.6 & 11.1 \\
kimi-k3 & \texttt{full} & 69.3 & 81.0 & 74.3 & 58.3 & 75.2 & 55.5 & 8.6 \\
claude-sonnet-5 & \texttt{search} & 69.0 & 83.8 & 73.6 & 56.6 & 75.2 & 54.5 & 13.3 \\
deepseek-v4-pro & \texttt{full} & 68.3 & 82.9 & 70.0 & 58.3 & 75.2 & 53.3 & 8.0 \\
deepseek-v4-pro & \texttt{search} & 68.1 & 83.8 & 72.1 & 55.4 & 73.1 & 53.8 & 12.5 \\
deepseek-v4-flash & \texttt{full} & 67.9 & 80.0 & 67.1 & 61.1 & 74.3 & 53.6 & 9.0 \\
qwen3.6-max & \texttt{search} & 66.9 & 81.0 & 59.3 & 64.6 & 71.7 & 48.6 & 11.4 \\
deepseek-v4-flash & \texttt{search} & 66.4 & 80.0 & 72.1 & 53.7 & 71.9 & 52.4 & 15.4 \\
qwen3.7-plus & \texttt{full} & 64.3 & 77.1 & 65.0 & 56.0 & 71.0 & 48.8 & 7.4 \\
glm-5.1 & \texttt{search} & 64.0 & 84.8 & 67.1 & 49.1 & 69.8 & 51.0 & 13.5 \\
glm-5.1 & \texttt{full} & 62.9 & 76.2 & 62.9 & 54.9 & 70.2 & 49.3 & 7.9 \\
qwen3.7-plus & \texttt{search} & 62.6 & 76.2 & 64.3 & 53.1 & 67.9 & 47.4 & 11.6 \\
qwen3.7-max & \texttt{search} & 62.4 & 73.3 & 60.7 & 57.1 & 68.1 & 45.2 & 10.9 \\
kimi-k2.6 & \texttt{search} & 61.4 & 77.1 & 69.3 & 45.7 & 61.4 & 46.9 & 11.8 \\
qwen3.7-max & \texttt{full} & 61.4 & 67.6 & 61.4 & 57.7 & 66.4 & 47.1 & 6.7 \\
qwen3.6-max & \texttt{full} & 61.0 & 74.3 & 60.0 & 53.7 & 65.7 & 46.4 & 6.4 \\
gpt-5.5 & \texttt{search} & 60.5 & 72.4 & 67.1 & 48.0 & 63.1 & 50.0 & 14.3 \\
qwen3.6-plus & \texttt{full} & 59.5 & 64.8 & 62.1 & 54.3 & 65.5 & 45.7 & 7.6 \\
kimi-k2.6 & \texttt{full} & 59.3 & 80.0 & 67.1 & 40.6 & 59.5 & 47.4 & 8.1 \\
qwen3.6-plus & \texttt{search} & 53.6 & 66.7 & 55.7 & 44.0 & 59.0 & 42.1 & 13.2 \\
\bottomrule
\end{tabular*}
\end{table*}

%% file: references.bib
@inproceedings{li-etal-2023-api,
  title = {{API}-Bank: A Comprehensive Benchmark for Tool-Augmented {LLM}s},
  author = {Li, Minghao and Zhao, Yingxiu and Yu, Bowen and Song, Feifan and Li, Hangyu and Yu, Haiyang and Li, Zhoujun and Huang, Fei and Li, Yongbin},
  editor = {Bouamor, Houda and Pino, Juan and Bali, Kalika},
  booktitle = {Proceedings of the 2023 Conference on Empirical Methods in Natural Language Processing},
  month = dec,
  year = {2023},
  address = {Singapore},
  publisher = {Association for Computational Linguistics},
  url = {https://aclanthology.org/2023.emnlp-main.187/},
  doi = {10.18653/v1/2023.emnlp-main.187},
  pages = {3102--3116}
}

@inproceedings{liu2024agentbench,
  title = {{AgentBench}: Evaluating {LLM}s as Agents},
  author = {Liu, Xiao and Yu, Hao and Zhang, Hanchen and Xu, Yifan and Lei, Xuanyu and Lai, Hanyu and Gu, Yu and Ding, Hangliang and Men, Kaiwen and Yang, Kejuan and Zhang, Shudan and Deng, Xiang and Zeng, Aohan and Du, Zhengxiao and Zhang, Chenhui and Shen, Sheng and Zhang, Tianjun and Su, Yu and Sun, Huan and Huang, Minlie and Dong, Yuxiao and Tang, Jie},
  booktitle = {The Twelfth International Conference on Learning Representations},
  year = {2024},
  url = {https://openreview.net/forum?id=zAdUB0aCTQ}
}

@inproceedings{zhou2024webarena,
  title = {{WebArena}: A Realistic Web Environment for Building Autonomous Agents},
  author = {Zhou, Shuyan and Xu, Frank F. and Zhu, Hao and Zhou, Xuhui and Lo, Robert and Sridhar, Abishek and Cheng, Xianyi and Ou, Tianyue and Bisk, Yonatan and Fried, Daniel and Alon, Uri and Neubig, Graham},
  booktitle = {The Twelfth International Conference on Learning Representations},
  year = {2024},
  url = {https://openreview.net/forum?id=rmiwIL98uQ}
}

@inproceedings{qin2024toolllm,
  title = {{ToolLLM}: Facilitating Large Language Models to Master 16000+ Real-world {API}s},
  author = {Qin, Yujia and Liang, Shihao and Ye, Yining and Zhu, Kunlun and Yan, Lan and Lu, Yaxi and Lin, Yankai and Cong, Xin and Tang, Xiangru and Qian, Bill and Zhao, Sihan and Hong, Lauren and Tian, Runchu and Xie, Ruobing and Zhou, Jie and Gerstein, Mark and Li, Dahai and Liu, Zhiyuan and Sun, Maosong},
  booktitle = {The Twelfth International Conference on Learning Representations},
  year = {2024},
  url = {https://openreview.net/forum?id=dHng2O0Jjr}
}

@inproceedings{shen2025shortcutsbench,
  title = {{ShortcutsBench}: A Large-Scale Real-world Benchmark for {API}-based Agents},
  author = {Shen, Haiyang and Li, Yue and Meng, Desong and Cai, Dongqi and Qi, Sheng and Zhang, Li and Xu, Mengwei and Ma, Yun},
  booktitle = {The Thirteenth International Conference on Learning Representations},
  year = {2025},
  url = {https://openreview.net/forum?id=c16a99558b0b4f6b10966ca9bdb98ade}
}

@inproceedings{elder2026liveapi,
  title = {{Live API-Bench}: 2500+ Live {API}s for Testing Multi-Step Tool Calling},
  author = {Elder, Benjamin and Murthi, Anupama and Kang, Jungkoo and Naik, Ankita Rajaram and Kate, Kiran and Basu, Kinjal and Contractor, Danish},
  booktitle = {Proceedings of the 19th Conference of the European Chapter of the Association for Computational Linguistics (Volume 1: Long Papers)},
  month = mar,
  year = {2026},
  address = {Rabat, Morocco},
  publisher = {Association for Computational Linguistics},
  pages = {3092--3124},
  doi = {10.18653/v1/2026.eacl-long.143},
  url = {https://aclanthology.org/2026.eacl-long.143/}
}

@article{pysklo2026agentdiff,
  title = {{Agent-Diff}: Benchmarking {LLM} Agents on Enterprise {API} Tasks via Code Execution with State-Diff-Based Evaluation},
  author = {Pysklo, Hubert M. and Zhuravel, Artem and Watson, Patrick D.},
  journal = {arXiv preprint arXiv:2602.11224},
  year = {2026},
  url = {https://arxiv.org/abs/2602.11224},
  doi = {10.48550/arXiv.2602.11224}
}

@article{lu2025agentrewardbench,
  title = {{AgentRewardBench}: Evaluating Automatic Evaluations of Web Agent Trajectories},
  author = {L{\`u}, Xing Han and Kazemnejad, Amirhossein and Meade, Nicholas and Patel, Arkil and Shin, Dongchan and Zambrano, Alejandra and Sta{\'n}czak, Karolina and Shaw, Peter and Pal, Christopher J. and Reddy, Siva},
  journal = {arXiv preprint arXiv:2504.08942},
  year = {2025},
  doi = {10.48550/arXiv.2504.08942},
  url = {https://arxiv.org/abs/2504.08942}
}

@inproceedings{xu2025contextualjudgebench,
  title = {Does Context Matter? {C}ontextual{J}udge{B}ench for Evaluating {LLM}-based Judges in Contextual Settings},
  author = {Xu, Austin and Bansal, Srijan and Ming, Yifei and Yavuz, Semih and Joty, Shafiq},
  booktitle = {Proceedings of the 63rd Annual Meeting of the Association for Computational Linguistics (Volume 1: Long Papers)},
  year = {2025},
  address = {Vienna, Austria},
  publisher = {Association for Computational Linguistics},
  pages = {9541--9564},
  doi = {10.18653/v1/2025.acl-long.470},
  url = {https://aclanthology.org/2025.acl-long.470/}
}

@inproceedings{styles2024workbench,
  title = {{WorkBench}: A Benchmark Dataset for Agents in a Realistic Workplace Setting},
  author = {Styles, Olly and Miller, Sam and Cerda-Mardini, Patricio and Guha, Tanaya and Sanchez, Victor and Vidgen, Bertie},
  booktitle = {First Conference on Language Modeling},
  year = {2024},
  url = {https://openreview.net/forum?id=4HNAwZFDcH}
}

@article{styles2026workbenchrevisited,
  title = {{WorkBench Revisited}: Workplace Agents Two Years On},
  author = {Styles, Olly and Miller, Sam},
  journal = {arXiv preprint arXiv:2606.13715},
  year = {2026},
  doi = {10.48550/arXiv.2606.13715},
  url = {https://arxiv.org/abs/2606.13715}
}

@inproceedings{trivedi2024appworld,
  title = {{AppWorld}: A Controllable World of Apps and People for Benchmarking Interactive Coding Agents},
  author = {Trivedi, Harsh and Khot, Tushar and Hartmann, Mareike and Manku, Ruskin and Dong, Vinty and Li, Edward and Gupta, Shashank and Sabharwal, Ashish and Balasubramanian, Niranjan},
  booktitle = {Proceedings of the 62nd Annual Meeting of the Association for Computational Linguistics (Volume 1: Long Papers)},
  year = {2024},
  address = {Bangkok, Thailand},
  publisher = {Association for Computational Linguistics},
  pages = {16022--16076},
  doi = {10.18653/v1/2024.acl-long.850},
  url = {https://aclanthology.org/2024.acl-long.850/}
}

@inproceedings{vishwakarma2025enterprisebench,
  title = {Can {LLM}s Help You at Work? A Sandbox for Evaluating {LLM} Agents in Enterprise Environments},
  author = {Vishwakarma, Harsh and Agarwal, Ankush and Patil, Ojas and Devaguptapu, Chaitanya and Chandran, Mahesh},
  booktitle = {Proceedings of the 2025 Conference on Empirical Methods in Natural Language Processing},
  year = {2025},
  address = {Suzhou, China},
  publisher = {Association for Computational Linguistics},
  pages = {9167--9201},
  doi = {10.18653/v1/2025.emnlp-main.466},
  url = {https://aclanthology.org/2025.emnlp-main.466/}
}

@inproceedings{huang2025crmarena,
  title = {{CRMArena}: Understanding the Capacity of {LLM} Agents to Perform Professional {CRM} Tasks in Realistic Environments},
  author = {Huang, Kung-Hsiang and Prabhakar, Akshara and Dhawan, Sidharth and Mao, Yixin and Wang, Huan and Savarese, Silvio and Xiong, Caiming and Laban, Philippe and Wu, Chien-Sheng},
  booktitle = {Proceedings of the 2025 Conference of the Nations of the Americas Chapter of the Association for Computational Linguistics: Human Language Technologies (Volume 1: Long Papers)},
  month = apr,
  year = {2025},
  address = {Albuquerque, New Mexico},
  publisher = {Association for Computational Linguistics},
  pages = {3830--3850},
  doi = {10.18653/v1/2025.naacl-long.194},
  url = {https://aclanthology.org/2025.naacl-long.194/}
}

@article{huang2026crmarena,
  title = {{CRMArena-Pro}: Holistic Assessment of {LLM} Agents Across Diverse Business Scenarios and Interactions},
  author = {Huang, Kung-Hsiang and Prabhakar, Akshara and Thorat, Onkar and Agarwal, Divyansh and Choubey, Prafulla Kumar and Mao, Yixin and Savarese, Silvio and Xiong, Caiming and Wu, Chien-Sheng},
  journal = {Transactions on Machine Learning Research},
  year = {2026},
  eprint = {2505.18878},
  archivePrefix = {arXiv},
  url = {https://openreview.net/forum?id=EPlpe3Fx1x}
}

@article{gupta2026worldworkflows,
  title = {World of Workflows: A Benchmark for Bringing World Models to Enterprise Systems},
  author = {Gupta, Lakshya and Li, Litao and Liu, Yizhe and Subramanian, Sriram Ganapathi and Suleman, Kaheer and Zhang, Zichen and Lu, Haoye and Pasupalak, Sumit},
  journal = {arXiv preprint arXiv:2601.22130},
  year = {2026},
  doi = {10.48550/arXiv.2601.22130},
  url = {https://arxiv.org/abs/2601.22130}
}

@article{malay2026enterpriseopsgym,
  title = {{EnterpriseOps-Gym}: Environments and Evaluations for Stateful Agentic Planning and Tool Use in Enterprise Settings},
  author = {Malay, Shiva Krishna Reddy and Nayak, Shravan and Nair, Jishnu Sethumadhavan and Davasam, Sagar and Tiwari, Aman and Madhusudhan, Sathwik Tejaswi and Nemala, Sridhar Krishna and Sunkara, Srinivas and Rajeswar, Sai},
  journal = {arXiv preprint arXiv:2603.13594},
  year = {2026},
  url = {https://arxiv.org/abs/2603.13594}
}

@article{zhong2026enterpriseclawbench,
  title = {{EnterpriseClawBench}: Benchmarking Agents from Real Workplace Sessions},
  author = {Zhong, Jincheng and Wang, Weizhi and Jiang, Che and Tian, Kai and Yuan, Zhenzhao and Yang, Junlin and Lei, Dianqiao and Zhang, Kaiyan},
  journal = {arXiv preprint arXiv:2606.23654},
  year = {2026},
  doi = {10.48550/arXiv.2606.23654},
  url = {https://arxiv.org/abs/2606.23654}
}

@article{mo2026entworld,
  title = {{EntWorld}: A Holistic Environment and Benchmark for Verifiable Enterprise {GUI} Agents},
  author = {Mo, Ying and Bai, Yu and Sun, Dapeng and Shi, Yuqian and Miao, Yukai and Chen, Li and Li, Dan},
  journal = {arXiv preprint arXiv:2601.17722},
  year = {2026},
  doi = {10.48550/arXiv.2601.17722},
  url = {https://arxiv.org/abs/2601.17722}
}

@article{agarwal2026enterpriselab,
  title = {{EnterpriseLab}: A Full-Stack Platform for Developing and Deploying Agents in Enterprises},
  author = {Agarwal, Ankush and Vishwakarma, Harsh and Nagaje, Suraj and Devaguptapu, Chaitanya},
  journal = {arXiv preprint arXiv:2603.21630},
  year = {2026},
  doi = {10.48550/arXiv.2603.21630},
  url = {https://arxiv.org/abs/2603.21630}
}

@article{ivanov2026anchor,
  title = {{Anchor}: Mitigating Artifact Drift in Agent Benchmark Generation},
  author = {Ivanov, Maksim and Rana, Abhijay},
  journal = {arXiv preprint arXiv:2605.26321},
  year = {2026},
  doi = {10.48550/arXiv.2605.26321},
  url = {https://arxiv.org/abs/2605.26321}
}

@article{shepard2026automationbench,
  title = {{AutomationBench}},
  author = {Shepard, Daniel and Salimans, Robin},
  journal = {arXiv preprint arXiv:2604.18934},
  year = {2026},
  url = {https://arxiv.org/abs/2604.18934}
}

@inproceedings{yao2024taubench,
  title = {{$\tau$-bench}: A Benchmark for Tool-Agent-User Interaction in Real-World Domains},
  author = {Yao, Shunyu and Shinn, Noah and Razavi, Pedram and Narasimhan, Karthik},
  booktitle = {The Thirteenth International Conference on Learning Representations},
  year = {2025}
}

@inproceedings{xie2024osworld,
  title = {{OSWorld}: Benchmarking Multimodal Agents for Open-Ended Tasks in Real Computer Environments},
  author = {Xie, Tianbao and Zhang, Danyang and Chen, Jixuan and Li, Xiaochuan and Zhao, Siheng and Cao, Ruisheng and Hua, Toh Jing and Cheng, Zhoujun and Shin, Dongchan and Lei, Fangyu and Liu, Yitao and Xu, Yiheng and Zhou, Shuyan and Savarese, Silvio and Xiong, Caiming and Zhong, Victor and Yu, Tao},
  booktitle = {Advances in Neural Information Processing Systems, Datasets and Benchmarks Track},
  volume = {37},
  pages = {52040--52094},
  publisher = {Curran Associates, Inc.},
  year = {2024},
  doi = {10.52202/079017-1650},
  url = {https://proceedings.neurips.cc/paper_files/paper/2024/hash/5d413e48f84dc61244b6be550f1cd8f5-Abstract-Datasets_and_Benchmarks_Track.html}
}

@inproceedings{mialon2024gaia,
  title = {{GAIA}: A Benchmark for General {AI} Assistants},
  author = {Mialon, Gr{\'e}goire and Fourrier, Cl{\'e}mentine and Swift, Craig and Wolf, Thomas and LeCun, Yann and Scialom, Thomas},
  booktitle = {The Twelfth International Conference on Learning Representations},
  year = {2024},
  url = {https://openreview.net/forum?id=fibxvahvs3}
}

@article{wang2024officebench,
  title = {{OfficeBench}: Benchmarking Language Agents across Multiple Applications for Office Automation},
  author = {Wang, Zilong and Cui, Yuedong and Zhong, Li and Zhang, Zimin and Yin, Da and Lin, Bill Yuchen and Shang, Jingbo},
  journal = {arXiv preprint arXiv:2407.19056},
  year = {2024},
  url = {https://arxiv.org/abs/2407.19056}
}

@article{wang2025odysseybench,
  title = {{OdysseyBench}: Evaluating {LLM} Agents on Long-Horizon Complex Office Application Workflows},
  author = {Wang, Weixuan and Han, Dongge and Diaz, Daniel Madrigal and Xu, Jin and R{\"u}hle, Victor and Rajmohan, Saravan},
  journal = {arXiv preprint arXiv:2508.09124},
  year = {2025},
  url = {https://arxiv.org/abs/2508.09124}
}

@article{tang2026workspacebench,
  title = {{Workspace-Bench} 1.0: Benchmarking {AI} Agents on Workspace Tasks with Large-Scale File Dependencies},
  author = {Tang, Zirui and Zhou, Xuanhe and Liu, Yumou and Li, Linchun and Wu, Yukai and Wang, Weizheng and Huang, Hongzhang and Zhou, Wei and Zhou, Jun and Song, Jiachen and Yu, Shaoli and Wang, Jinqi and Zhou, Zihang and Zhou, Hongyi and Lv, Yuting and Li, Jinyang and Liu, Jiashuo and Chen, Ruoyu and Liu, Chunwei and Li, GuoLiang and Kang, Jihua and Wu, Fan},
  journal = {arXiv preprint arXiv:2605.03596},
  year = {2026},
  url = {https://arxiv.org/abs/2605.03596}
}

@inproceedings{xu2025agentcompany,
  title={TheAgentCompany: Benchmarking LLM Agents on Consequential Real World Tasks},
  author={Xu, Frank Fangzheng and Song, Yufan and Li, Boxuan and Tang, Yuxuan and Jain, Kritanjali and Bao, Mengxue and Wang, Zora and Zhou, Xuhui and Guo, Zhitong and Cao, Murong and Yang, Mingyang and Lu, Hao Yang and Martin, Amaad and Su, Zhe and Maben, Leander and Mehta, Raj and Chi, Wayne and Jang, Lawrence and Xie, Yiqing and Zhou, Shuyan and Neubig, Graham},
  booktitle={Advances in Neural Information Processing Systems, Datasets and Benchmarks Track},
  volume={38},
  publisher={Curran Associates, Inc.},
  year={2025},
  url={https://proceedings.neurips.cc/paper_files/paper/2025/hash/0d744742f6fac4d1134c019b7cef3c8a-Abstract-Datasets_and_Benchmarks_Track.html}
}

@inproceedings{wang2024gta,
  title={GTA: A Benchmark for General Tool Agents},
  author={Wang, Jize and Ma, Zerun and Li, Yining and Zhang, Songyang and Chen, Cailian and Chen, Kai and Le, Xinyi},
  booktitle={Advances in Neural Information Processing Systems, Datasets and Benchmarks Track},
  volume={37},
  pages={75749--75790},
  publisher={Curran Associates, Inc.},
  doi={10.52202/079017-2412},
  year={2024},
  url={https://proceedings.neurips.cc/paper_files/paper/2024/hash/8a75ee6d4b2eb0b777f549a32a5a5c28-Abstract-Datasets_and_Benchmarks_Track.html}
}

@inproceedings{mcgregor2025benchrisk,
  title = {Risk Management for Mitigating Benchmark Failure Modes: {BenchRisk}},
  author = {McGregor, Sean and Lu, Victor and Tashev, Vassil and Foundjem, Armstrong and Ramasethu, Aishwarya and Kazemi Zarkouei, Sadegh AlMahdi and Knotz, Chris and Chen, Kongtao and Parrish, Alicia and Reuel, Anka and Frase, Heather},
  booktitle = {Advances in Neural Information Processing Systems, Datasets and Benchmarks Track},
  volume = {38},
  publisher = {Curran Associates, Inc.},
  year = {2025},
  url = {https://proceedings.neurips.cc/paper_files/paper/2025/hash/92a0af72659802465884eaad8443ea89-Abstract-Datasets_and_Benchmarks_Track.html}
}

@inproceedings{bean2025constructvalidity,
  title = {Measuring What Matters: Construct Validity in Large Language Model Benchmarks},
  author = {Bean, Andrew M. and Kearns, Ryan Othniel and Romanou, Angelika and Hafner, Franziska Sofia and Mayne, Harry and Batzner, Jan and Foroutan Eghlidi, Negar and Schmitz, Chris and Korgul, Karolina and Batra, Hunar and Deb, Oishi and Beharry, Emma and Emde, Cornelius and Foster, Thomas and Gausen, Anna and Grandury, Mar{\'i}a and Han, Simeng and Hofmann, Valentin and Ibrahim, Lujain and Kim, Hazel and Kirk, Hannah Rose and Lin, Fangru and Liu, Gabrielle Kaili-May and Luettgau, Lennart and Magomere, Jabez and Rystr{\o}m, Jonathan and Sotnikova, Anna and Yang, Yushi and Zhao, Yilun and Bibi, Adel and Bosselut, Antoine and Clark, Ronald and Cohan, Arman and Foerster, Jakob and Gal, Yarin and Hale, Scott A. and Raji, Inioluwa Deborah and Summerfield, Christopher and Torr, Philip H. S. and Ududec, Cozmin and Rocher, Luc and Mahdi, Adam},
  booktitle = {Advances in Neural Information Processing Systems, Datasets and Benchmarks Track},
  volume = {38},
  publisher = {Curran Associates, Inc.},
  year = {2025},
  url = {https://proceedings.neurips.cc/paper_files/paper/2025/hash/1967e0fc3aa6cbbace562f5cb8e3954e-Abstract-Datasets_and_Benchmarks_Track.html}
}

@inproceedings{koh2024visualwebarena,
  title = {{VisualWebArena}: Evaluating Multimodal Agents on Realistic Visual Web Tasks},
  author = {Koh, Jing Yu and Lo, Robert and Jang, Lawrence and Duvvur, Vikram and Lim, Ming and Huang, Po-Yu and Neubig, Graham and Zhou, Shuyan and Salakhutdinov, Russ and Fried, Daniel},
  booktitle = {Proceedings of the 62nd Annual Meeting of the Association for Computational Linguistics (Volume 1: Long Papers)},
  month = aug,
  year = {2024},
  address = {Bangkok, Thailand},
  publisher = {Association for Computational Linguistics},
  pages = {881--905},
  doi = {10.18653/v1/2024.acl-long.50},
  url = {https://aclanthology.org/2024.acl-long.50/}
}

@inproceedings{ma2024agentboard,
  title = {{AgentBoard}: An Analytical Evaluation Board of Multi-turn {LLM} Agents},
  author = {Ma, Chang and Zhang, Junlei and Zhu, Zhihao and Yang, Cheng and Yang, Yujiu and Jin, Yaohui and Lan, Zhenzhong and Kong, Lingpeng and He, Junxian},
  booktitle = {Advances in Neural Information Processing Systems, Datasets and Benchmarks Track},
  volume = {37},
  publisher = {Curran Associates, Inc.},
  year = {2024},
  doi = {10.52202/079017-2365},
  url = {https://proceedings.neurips.cc/paper_files/paper/2024/hash/877b40688e330a0e2a3fc24084208dfa-Abstract-Datasets_and_Benchmarks_Track.html}
}

@inproceedings{rawles2025androidworld,
  title = {{AndroidWorld}: A Dynamic Benchmarking Environment for Autonomous Agents},
  author = {Rawles, Christopher and Clinckemaillie, Sarah and Chang, Yifan and Waltz, Jonathan and Lau, Gabrielle and Fair, Marybeth and Li, Alice and Bishop, William E. and Li, Wei and Campbell-Ajala, Folawiyo and Toyama, Daniel Kenji and Berry, Robert James and Tyamagundlu, Divya and Lillicrap, Timothy P. and Riva, Oriana},
  booktitle = {The Thirteenth International Conference on Learning Representations},
  year = {2025},
  url = {https://openreview.net/forum?id=il5yUQsrjC}
}

@inproceedings{drouin2024workarena,
  title = {{WorkArena}: How Capable are Web Agents at Solving Common Knowledge Work Tasks?},
  author = {Drouin, Alexandre and Gasse, Maxime and Caccia, Massimo and Laradji, Issam H. and Del Verme, Manuel and Marty, Tom and Vazquez, David and Chapados, Nicolas and Lacoste, Alexandre},
  booktitle = {Proceedings of the 41st International Conference on Machine Learning},
  pages = {11642--11662},
  year = {2024},
  volume = {235},
  series = {Proceedings of Machine Learning Research},
  publisher = {PMLR},
  url = {https://proceedings.mlr.press/v235/drouin24a.html}
}

@inproceedings{boisvert2024workarenaplus,
  title = {{WorkArena++}: Towards Compositional Planning and Reasoning-based Common Knowledge Work Tasks},
  author = {Boisvert, L{\'e}o and Thakkar, Megh and Gasse, Maxime and Caccia, Massimo and Le Sellier De Chezelles, Thibault and Cappart, Quentin and Chapados, Nicolas and Lacoste, Alexandre and Drouin, Alexandre},
  booktitle = {Advances in Neural Information Processing Systems, Datasets and Benchmarks Track},
  volume = {37},
  publisher = {Curran Associates, Inc.},
  year = {2024},
  doi = {10.52202/079017-0195},
  url = {https://proceedings.neurips.cc/paper_files/paper/2024/hash/0b82662b6c32e887bb252a74d8cb2d5e-Abstract-Datasets_and_Benchmarks_Track.html}
}

@inproceedings{guo2024stabletoolbench,
  title = {{StableToolBench}: Towards Stable Large-Scale Benchmarking on Tool Learning of Large Language Models},
  author = {Guo, Zhicheng and Cheng, Sijie and Wang, Hao and Liang, Shihao and Qin, Yujia and Li, Peng and Liu, Zhiyuan and Sun, Maosong and Liu, Yang},
  booktitle = {Findings of the Association for Computational Linguistics: ACL 2024},
  month = aug,
  year = {2024},
  address = {Bangkok, Thailand},
  publisher = {Association for Computational Linguistics},
  pages = {11143--11156},
  doi = {10.18653/v1/2024.findings-acl.664},
  url = {https://aclanthology.org/2024.findings-acl.664/}
}

@inproceedings{huang2024ultratool,
  title = {Planning, Creation, Usage: Benchmarking {LLM}s for Comprehensive Tool Utilization in Real-World Complex Scenarios},
  author = {Huang, Shijue and Zhong, Wanjun and Lu, Jianqiao and Zhu, Qi and Gao, Jiahui and Liu, Weiwen and Hou, Yutai and Zeng, Xingshan and Wang, Yasheng and Shang, Lifeng and Jiang, Xin and Xu, Ruifeng and Liu, Qun},
  booktitle = {Findings of the Association for Computational Linguistics: ACL 2024},
  month = aug,
  year = {2024},
  address = {Bangkok, Thailand},
  publisher = {Association for Computational Linguistics},
  pages = {4363--4400},
  doi = {10.18653/v1/2024.findings-acl.259},
  url = {https://aclanthology.org/2024.findings-acl.259/}
}

@inproceedings{lu2025toolsandbox,
  title = {{ToolSandbox}: A Stateful, Conversational, Interactive Evaluation Benchmark for {LLM} Tool Use Capabilities},
  author = {Lu, Jiarui and Holleis, Thomas and Zhang, Yizhe and Aumayer, Bernhard and Nan, Feng and Bai, Haoping and Ma, Shuang and Ma, Shen and Li, Mengyu and Yin, Guoli and Wang, Zirui and Pang, Ruoming},
  booktitle = {Findings of the Association for Computational Linguistics: NAACL 2025},
  month = apr,
  year = {2025},
  address = {Albuquerque, New Mexico},
  publisher = {Association for Computational Linguistics},
  pages = {1160--1183},
  doi = {10.18653/v1/2025.findings-naacl.65},
  url = {https://aclanthology.org/2025.findings-naacl.65/}
}

@inproceedings{patil2025bfcl,
  title = {The Berkeley Function Calling Leaderboard ({BFCL}): From Tool Use to Agentic Evaluation of Large Language Models},
  author = {Patil, Shishir G. and Mao, Huanzhi and Yan, Fanjia and Ji, Charlie Cheng-Jie and Suresh, Vishnu and Stoica, Ion and Gonzalez, Joseph E.},
  booktitle = {Proceedings of the 42nd International Conference on Machine Learning},
  pages = {48371--48392},
  year = {2025},
  volume = {267},
  series = {Proceedings of Machine Learning Research},
  publisher = {PMLR},
  url = {https://proceedings.mlr.press/v267/patil25a.html}
}

@inproceedings{wang2024mint,
  title = {{MINT}: Evaluating {LLM}s in Multi-turn Interaction with Tools and Language Feedback},
  author = {Wang, Xingyao and Wang, Zihan and Liu, Jiateng and Chen, Yangyi and Yuan, Lifan and Peng, Hao and Ji, Heng},
  booktitle = {The Twelfth International Conference on Learning Representations},
  year = {2024},
  url = {https://openreview.net/forum?id=jp3gWrMuIZ}
}

@inproceedings{zheng2023judging,
  title = {Judging {LLM}-as-a-Judge with {MT}-Bench and Chatbot Arena},
  author = {Zheng, Lianmin and Chiang, Wei-Lin and Sheng, Ying and Zhuang, Siyuan and Wu, Zhanghao and Zhuang, Yonghao and Lin, Zi and Li, Zhuohan and Li, Dacheng and Xing, Eric P. and Zhang, Hao and Gonzalez, Joseph E. and Stoica, Ion},
  booktitle = {Advances in Neural Information Processing Systems, Datasets and Benchmarks Track},
  volume = {36},
  pages = {46595--46623},
  publisher = {Curran Associates, Inc.},
  year = {2023},
  url = {https://proceedings.neurips.cc/paper_files/paper/2023/hash/91f18a1287b398d378ef22505bf41832-Abstract-Datasets_and_Benchmarks.html}
}

@inproceedings{raina2024judgerobust,
  title = {Is {LLM}-as-a-Judge Robust? Investigating Universal Adversarial Attacks on Zero-shot {LLM} Assessment},
  author = {Raina, Vyas and Liusie, Adian and Gales, Mark},
  booktitle = {Proceedings of the 2024 Conference on Empirical Methods in Natural Language Processing},
  month = nov,
  year = {2024},
  address = {Miami, Florida, USA},
  publisher = {Association for Computational Linguistics},
  pages = {7499--7517},
  doi = {10.18653/v1/2024.emnlp-main.427},
  url = {https://aclanthology.org/2024.emnlp-main.427/}
}
